\documentclass{article}

\usepackage{microtype}
\usepackage{graphicx}
\usepackage{subfigure}
\usepackage{tabularx,booktabs,makecell}
\usepackage{makecell}   % optional, for better line breaks in headers
\usepackage{multirow}
\usepackage{graphicx}
\usepackage[table,dvipsnames]{xcolor}
\usepackage{arydshln}

\usepackage[most]{tcolorbox}   % 核心包，提供 aptcolorbox 环境
\usepackage{caption}           % 允许 \captionof{table} 等命令
\usepackage{xcolor}            % 支持自定义颜色 (tcolorbox 需要)
\usepackage{verbatim}          % 为了安全使用 \begin{verbatim}

\usepackage{hyperref}
\usepackage{siunitx}
\definecolor{toolcall}{RGB}{180, 130, 130}
\definecolor{toolresponse}{RGB}{70,130,85}
\definecolor{answer}{RGB}{175,50,50}
\definecolor{think}{RGB}{100,70,175}

\newcommand{\boldtt}[1]{{\fontseries{b}\selectfont\texttt{#1}}}
\newcommand{\toolcall}[1]{\textcolor{toolcall}{\boldtt{<tool\_call>#1</tool\_call>}}}
\newcommand{\toolresponse}[1]{\textcolor{toolresponse}{\boldtt{<tool\_response>#1</tool\_response>}}}
\newcommand{\answer}[1]{\textcolor{answer}{\boldtt{<answer>#1</answer>}}}
\newcommand{\think}[1]{\textcolor{think}{\boldtt{<think>#1</think>}}}
\newcommand{\thought}{\textcolor{think}{\boldtt{Thought}}}
\newcommand{\action}{\textcolor{toolcall}{\boldtt{Action}}}
\newcommand{\observation}{\textcolor{toolresponse}{\boldtt{Observation}}}
\newcommand{\singlethink}{\textcolor{think}{\boldtt{<think>}}}
\newcommand{\singletoolcall}{\textcolor{toolcall}{\boldtt{<tool\_call>}}}
\newcommand{\singletoolresponse}{\textcolor{toolresponse}{\boldtt{<tool\_response>}}}
\newcommand{\singleanswer}{\textcolor{answer}{\boldtt{<answer>}}}

\newcommand{\reason}[1]{\textcolor{think} {\texttt{<think>}} #1 \textcolor{think}{\texttt{</think>}}}
\newcommand{\call}[1]{\textcolor{toolcall} {\texttt{<tool\_call>}} #1 \textcolor{toolcall}{\texttt{</tool\_call>}}}
\newcommand{\response}[1]{\textcolor{toolresponse}{\texttt{<tool\_response>}} #1 \textcolor{toolresponse}{\texttt{</tool\_response>}}}

\newcommand{\ans}[1]{\textcolor{answer}{\texttt{<answer>}} #1 \textcolor{answer}{\texttt{</answer>}}}

\definecolor{algcommentblue}{RGB}{0,102,153}

\newcommand{\algcomment}[1]{%
  \STATE \textcolor{algcommentblue}{\textit{// #1}}%
}

\newcommand{\framework}{SynPlanResearch-R1}
\definecolor{lightblue}{RGB}{230,240,255}
\definecolor{groupgray}{RGB}{245,245,245}

\usepackage{icml2025}

\makeatletter
\gdef\isaccepted{1}
\renewcommand{\Notice@String}{}
\makeatother

\usepackage{amsmath}
\usepackage{amssymb}
\usepackage{mathtools}
\usepackage{amsthm}

\usepackage[capitalize,noabbrev]{cleveref}

\theoremstyle{plain}

\theoremstyle{definition}

\theoremstyle{remark}

\usepackage[textsize=tiny]{todonotes}

\icmltitlerunning{Submission and Formatting Instructions for ICML 2025}

\begin{document}

\twocolumn[
\icmltitle{SynPlanResearch-R1: Encouraging Tool Exploration for Deep Research with Synthetic Plans}

% It is OKAY to include author information, even for blind
% submissions: the style file will automatically remove it for you
% unless you've provided the [accepted] option to the icml2025
% package.

% List of affiliations: The first argument should be a (short)
% identifier you will use later to specify author affiliations
% Academic affiliations should list Department, University, City, Region, Country
% Industry affiliations should list Company, City, Region, Country

% You can specify symbols, otherwise they are numbered in order.
% Ideally, you should not use this facility. Affiliations will be numbered
% in order of appearance and this is the preferred way.
\icmlsetsymbol{equal}{*}

\begin{icmlauthorlist}
\icmlauthor{Hansi Zeng}{umass}
\icmlauthor{Zoey Li}{amazon}
\icmlauthor{Yifan Gao}{amazon}
\icmlauthor{Chenwei Zhang}{amazon}
\icmlauthor{Xiaoman Pan}{amazon}
\icmlauthor{Tao Yang}{amazon}
\icmlauthor{Fengran Mo}{montreal}
\icmlauthor{Jiacheng Lin}{uiuc}
\icmlauthor{Xian Li}{amazon}
\icmlauthor{Jingbo Shang}{ucsd}
% \icmlauthor{}{sch}
% \icmlauthor{Firstname8 Lastname8}{sch}
% \icmlauthor{Firstname8 Lastname8}{yyy,comp}
% \icmlauthor{}{sch}
% \icmlauthor{}{sch}
\end{icmlauthorlist}

\icmlaffiliation{umass}{University of Massachusetts Amherst }
\icmlaffiliation{amazon}{Amazon}
\icmlaffiliation{montreal}{University of Montreal}
\icmlaffiliation{ucsd}{University of San Diego}
\icmlaffiliation{uiuc}{University of Illinois Urbana-Champaign}

\icmlcorrespondingauthor{Hansi Zeng}{hzeng@cs.umass.edu}
% \icmlcorrespondingauthor{Firstname2 Lastname2}{first2.last2@www.uk}

% You may provide any keywords that you
% find helpful for describing your paper; these are used to populate
% the "keywords" metadata in the PDF but will not be shown in the document
\icmlkeywords{Machine Learning, ICML}

\vskip 0.3in
]

% this must go after the closing bracket ] following \twocolumn[ ...

% This command actually creates the footnote in the first column
% listing the affiliations and the copyright notice.
% The command takes one argument, which is text to display at the start of the footnote.
% The \icmlEqualContribution command is standard text for equal contribution.
% Remove it (just {}) if you do not need this facility.

%\printAffiliationsAndNotice{}  % leave blank if no need to mention equal contribution
\printAffiliationsAndNotice{\icmlEqualContribution} % otherwise use the standard text.

\begin{abstract}
Research Agents\footnote{Research Agents can either be text-only or GUI-based. In this paper, we focus on the text-only setting. } enable models to gather information from the web using tools to answer user queries, requiring them to dynamically interleave internal reasoning with tool use. While such capabilities can in principle be learned via reinforcement learning with verifiable rewards (RLVR), we observe that agents often exhibit poor exploration behaviors, including premature termination and biased tool usage. As a result, RLVR alone yields limited improvements. We propose \framework, a framework that synthesizes tool-use trajectories that encourage deeper exploration to shape exploration during cold-start supervised fine-tuning, providing a strong initialization for subsequent RL. Across seven multi-hop and open-web benchmarks, \framework improves performance by up to 6.0\% on Qwen3-8B and 5.8\% on Qwen3-4B backbones respectively compared to SOTA baselines. Further analyses of tool-use patterns and training dynamics compared to baselines shed light on the factors underlying these gains. Our code is publicly available at \url{https://github.com/HansiZeng/syn-plan-research}.
\end{abstract}

\section{Introduction}
% \zoey{Current logic of intro follows: introduce research agents -- early prompting and SFT training for research agents -- RLVR training. I think we should emphasize more on tool use and why diverse tools can expand agent capability.}

Large Language Models (LLMs) have demonstrated remarkable capabilities in complex reasoning, decision-making, and interacting with external tools~\cite{ragen,retool,toRL,toolstar}. This has driven the rise of research agents, which can autonomously explore the web, retrieve and synthesize information from diverse sources, and generate comprehensive answers or reports to fulfill complex, knowledge-intensive user queries~\cite{webgpt,search-r1,webthinker,arpo,gigpo}.

The precursor of ResearchAgents is retrieval-augmented generation (RAG)~\cite{rag,realm}, with early systems directly feeding document embeddings into the generation process and later systems incorporating the retrieved documents into the context. 
% Systems ranged from single-hop retrieval to prompt-engineered, multi-step interleavings of reasoning and retrieval~\cite{shi-etal-2024-generate,jin-etal-2024-bider,guan2025deepragthinkingretrievestep,selfrag,ir-cot}. 
A fundamental challenge was determining when extra context was needed, as it was observed that low-quality retrieval could often distract the model and degrade efficiency as well. 
% A fundamental challenge with these prompt-based approaches was the lack of explicit supervision for the intermediate processes of reasoning and reflection, limiting their effectiveness. 
Self-RAG~\cite{selfrag} sought to address this by training the model to reflect on the necessity and relevance of retrieved passages with rewards distilled from powerful models such as GPT-4~\cite{GPT-4}.
%to teach the agent useful actions—such as when to retrieve, generate, or critique evidence. 
However, this method still operates within a very limited action space: the model only has the option to decide whether to use retrieval in a single-turn interaction, without control over what to read or when to stop. Fast-forward to today, Research Agents (or Deep Research systems) run dozens of rounds of search queries, paired with document and even image processing tools, to tackle complex user requests.

% However, such supervision is inherently off-policy: the agent imitates a teacher’s traces rather than learning from its own interactions. Lacking on-policy experience and credit assignment over its own decisions, the agent tends to inherit the teacher’s ceiling and cannot reach task-optimal performance.
How can we effectively train Research Agents with multi-turn tool use? 
Apart from imitation learning using expert trajectories, Reinforcement Learning from Verifiable Rewards (RLVR)~\cite{deepseek2025deepseekr1,shao2024deepseekmath, lambert2024tulu3} presents a more direct alternative. This on-policy paradigm optimizes the research agent using end-task reward alone (whether the final answer contains the ground truth or not), eliminating the need to train a dense reward model. 
The process begins by prompting the agent with tool descriptions and desired behaviors, often starting from a warm-start policy obtained via cold-start supervised finetuning (SFT) that teaches basic tool usage and formatting~\cite{search-r1,webthinker,arpo,toolstar}. The policy is then refined solely through a scalar outcome-based reward. 
% This elegant approach allows the agent to self-improve through its own experience, requiring no engineered intermediate rewards. By directly optimizing for the final goal, it empowers the agent to discover a performant, task-specific policy, free from the constraints of imitation.
% \zoey{Probably need to further cut down this paragraph}
This elegant approach enables the agent to self-improve through its own experience, ultimately converging to a performant, task-specific policy.

% However, because RLVR is on-policy and bootstraps from its own rollouts, what the learner can improve is strictly limited by what its current policy visits; a weak initial policy confines the agent to a narrow band of suboptimal behaviors from the start~\cite{interp-policy-gradient}.
% In research agents, this limitation manifests itself practically: the agent frequently concludes reasoning with insufficient tool calls, delivering answers prematurely. Moreover, when multiple tools are available, the agent demonstrates a clear bias towards familiar patterns.
%—heavily favoring the primary web\_search tool while underutilizing complementary tools like click\_url for deeper page exploration.
However, we found that RLVR-trained web research agents often fail to discover effective tool-use trajectories for solving complex queries. 
In practice, this failure typically manifests in two ways: (1) the agent terminates reasoning prematurely after issuing too few search queries, and (2) it struggles to effectively compose multiple tools within a single reasoning chain, resulting in shallow or fragmented evidence gathering.
We attribute this failure to the on-policy nature of RLVR~\cite{interp-policy-gradient}: because learning bootstraps from the agent’s own rollouts, a weak initialization restricts exploration and traps the policy in suboptimal tool-use behaviors.

% To address this initialization bottleneck, we propose \framework, a framework that synthesizes diverse, complementary tool-use trajectories for the cold-start SFT phase. We adopt the ReAct thought–action–observation cycle and expand exploration through three steps. First, we create a randomized tool plan—variable-length, starting with web\_search and then stochastically interleaving actions such as click\_url. We attach this plan to the query, prompting an large reasoning model (LRM) to generate trajectories with broader tool use. 
% Second, because vanilla LRMs often deviate from the plan, we add a tool-dependent cue at the start of each thought to softly steer the next action while leaving the reasoning content unconstrained. Third, for each query we produce multiple trajectories under distinct plans and retain those that pass answer-correctness and format checks, balancing diversity with reliability. The resulting traces prime SFT with a stronger exploration prior, providing a much more capable foundation for subsequent RL refinement.
To address the initialization bottleneck, we propose \framework, a plan-guided data synthesis framework for cold-start SFT prior to RL. The key idea is to synthesize tool-use trajectories that encourage deeper exploration while producing correct and well-formed outcomes, and to use these trajectories to shape a stronger initial policy that raises the ceiling of subsequent RL optimization.
Built upon the ReAct framework~\cite{react}, we generate lightweight, randomized tool plans to guide trajectory generation, prompting a large reasoning model to produce longer reasoning chains with richer tool interactions. To encourage plan adherence without stifling model reasoning, we introduce simple tool-dependent cues that softly steer action selection, and retain only trajectories that are answer-correct and free of format errors.
During the RL phase, we propose two simple yet effective stabilization strategies to address challenges in multi-turn tool-based training, including masking the policy loss for invalid trajectories that exceed turn or token limits, and terminating generation immediately when tool calls violate the required JSON schema, rather than continuing generation via reflection.

% \zoey{Although not our main contribution, I think we can discuss a bit on the difficulty of RL training and the tricks needed.}

% We evaluate \framework across seven challenging benchmarks spanning multi-hop QA (HotpotQA~\cite{hotpotqa} 2WikiMultihopQA~\cite{2wiki}, MuSiQue~\cite{musique}, Bamboogle~\cite{bamboogle}) and harder web-based tasks (GPQA~\cite{gpqa}, WebWalkerQA~\cite{webwalker}, GAIA~\cite{gaia}). Experimental results demonstrate that our framework consistently outperforms strong control baselines, including those using naive cold-start SFT followed by RL and RL-from-scratch. A key finding is the strong correlation between performance and exploration depth: accuracy improves monotonically with the number of tool calls, with our method consistently occupying the upper-right envelope—achieving both the highest tool usage and best scores. Further analysis of training dynamics reveals that our approach maintains higher policy entropy early in training, reflecting broader exploration, and ultimately surpasses all baselines in reward as RL effectively converts this exploration into superior solutions. 
% Against prior state-of-the-art research agents, \framework delivers consistent gains across model scales. On 8B models, it achieves $5.1\%$ improvement on multi-hop QA and nearly a $8.7\%$ improvement on the more challenging advanced QA benchmarks. The 4B models follow the same trend, showing improvements of about $5.2\%$ on multi-hop QA and $6.0\%$ on advanced QA tasks.

We evaluate \framework across seven challenging benchmarks spanning multi-hop question answering and open-web research tasks ( (HotpotQA~\cite{hotpotqa} 2WikiMultihopQA~\cite{2wiki}, MuSiQue~\cite{musique}, Bamboogle~\cite{bamboogle}, GPQA~\cite{gpqa}, WebWalkerQA~\cite{webwalker}, GAIA~\cite{gaia}). 
Compared to prior state-of-the-art research agents, \framework delivers consistent gains across model scales, achieving improvements of up to $5.1\%$ on multi-hop QA and $8.7\%$ on advanced QA benchmarks with 8B models, and $5.2\%$ and $6.0\%$ gains with 4B models.
Beyond overall performance, we conduct analysis against control baselines (e.g., vanilla RL from cold-start SFT and RL-from-scratch) and find that \framework enables deeper tool-use exploration and maintains higher policy entropy during early training, which together explain its performance advantages.
% By shaping exploration at the cold-start stage, \framework provides a stronger initializaiton for on-policy reinforcement learning and yields reliable improvements in complex reasoning and search settings.~\hansi{make it more concise}
%
% These results collectively establish \framework as a stronger and more reliable research agent across diverse reasoning and search settings.

\section{Related Works}
\subsection{Reinforcement Learning with Verifiable Reward} 
Reinforcement Learning with Verifiable Rewards (RLVR)~\cite{lambert2024tulu3,kaufmann2025survey} has emerged as a key paradigm for optimizing large language models (LLMs) in domains such as mathematical reasoning and code generation~\cite{openai2024learning, deepseek2025deepseekr1, shao2024deepseekmath, kaufmann2025survey}. This approach was assumed to be the powering force behind OpenAI’s o1 system~\cite{openai2024learning} and subsequently advanced by models such as DeepSeek-R1~\cite{deepseek2025deepseekr1}, Kimi K1.5~\cite{kimi2025kimi}, and QwQ~\cite{qwen2024qwq}
% which demonstrated the scalability and effectiveness of outcome-based reinforcement learning. 
Recently, value-network free optimization methods (GRPO~\cite{shao2024deepseekmath}, RLOO~\cite{ahmadian2024back}) have gained the spotlight due to their effectiveness (as demonstrated through DeepSeek-R1) and simplicity compared to established methods such as PPO~\cite{schulman2017proximalpolicyoptimizationalgorithms}. Various enhancements have been proposed to enhance training stability and efficiency, including DAPO~\cite{yu2025dapoopensourcellmreinforcement}, GSPO~\cite{zheng2025group}, and
IS correction techniques \cite{yao2025your, liu2025when}.
%other variants of policy-gradient refinement~\cite{hu2025reinforceplusplus, yao2025your, liu2025when}. 
%the de facto optimization algorithm, yet research on enhancing RLVR’s overall stability and efficiency continues to evolve along two main directions. The first, an algorithmic perspective, focuses on developing improved policy objectives and optimization schemes, including DAPO~\cite{yu2025dapoopensourcellmreinforcement}, GSPO~\cite{zheng2025group}, and other variants of policy-gradient refinement~\cite{ahmadian2024back, hu2025reinforceplusplus, yao2025your, liu2025when}. 
% The second, a numerical and infrastructural perspective, investigates the root causes of instability—such as discrepancies between training and inference behaviors~\cite{liu2025when, yao2025your, team2025every}—and proposes system-level remedies, including precision control (e.g., FP16 vs. BF16) and distributed optimization techniques~\cite{micikevicius2017mixed, kalamkar2019study, sheng2024hybridflow, kwon2023efficient, zhao2023pytorch}. 
Although RLVR has seen wide success in improving model performance, it is still debatable whether RLVR genuinely teaches the model new skills or simply sharpens existing ones~\cite{yue2025limit-of-rlvr,liu2025prorl}. This connects to our observation that RLVR alone cannot make the model escape from its own biases to explore more effective solutions.
% Complementary to these engineering advances, a parallel research line seeks to uncover the underlying mechanisms of RLVR~\cite{cui2025entropy, li2025llms, wang2025beyond}. Studies in this direction reveal the critical influence of high-entropy tokens during exploration~\cite{wang2025beyond, vassoyan2025ignore, cui2025entropy}, the model’s surprising ability to learn from sparse, single-sample rewards~\cite{wang2025reinforcement}, and its inherent tendency to overfit to reward structure rather than genuine semantic correctness~\cite{li2025llms, gandhi2025cognitive}.

\subsection{Research Agents}
Answering knowledge-intensive questions require interaction with a knowledge base or the open web. This idea led to the development of retrieval-augmented language models~\cite{rag,pmlr-retro,selfrag,guan2025deepragthinkingretrievestep,ir-cot} which are either trained with an extended language modeling objective or uses non-intrusive multi-step prompting pipeline. 
%with rule-driven multi-step pipelines, but these workflows depend on handcrafted heuristics and often lack learning signals for optimizing intermediate decisions.
Deep research agents~\cite{openai2025deepresearch,google2025geminideepresearch} extend large reasoning models (LRMs) by coupling them with external tools—most prominently web search and browsing—to gather, verify, and synthesize evidence from the open web for complex, knowledge-intensive queries
%~\cite{search-r1,song2025r1searcher,search-o1}.
% \zoey{Given that WebGPT was trained with PPO/rejection sampling in 2021, I don't think we should claim RLVR as a key novelty. Reworded the below paragraph.}
% Early systems grounded this idea in  
While dense retrieval can be jointly trained with the LM, web-browsing behavior is non-differentiable.  
The recent surge of RLVR has made training research agents more accessible~\cite{search-r1,ragen}, without the need to collect human judgements for reward model training as in \cite{webgpt}.
%Recently, RLVR has emerged as a promising paradigm for building research agents. It directly optimizes the entire agent—spanning reasoning, retrieval, and tool use behaviors, to maximize specific reward such as answer correctness~\cite{search-r1,ragen}. 
% This approach enables joint improvement of both the reasoning policy and the decision process for invoking tools. 
Current research in this direction focuses on three complementary aspects: (1) developing more effective optimization algorithms tailored to web-based research~\cite{arpo,search-r1,dong2025aepo}, (2) constructing high-quality training datasets
for both cold-start supervised fine-tuning and subsequent reinforcement learning
%—either by synthesizing challenging queries or curating reliable examples from existing sources
~\cite{simpledeepsearcher,webthinker,tao2025webshaperagenticallydatasynthesizing,li2025websailornavigatingsuperhumanreasoning,liu2025webexplorer}, and (3) developing more efficient RL frameworks to support long horizon training~\cite{gao2025beyondTT}.
%(3) integrating web search with auxiliary tools such as code interpreters for reasoning or summarizers for compressing and filtering long web results to enhance the agent’s overall reasoning depth and efficiency~\cite{webthinker,toolstar, qin2024toolllm, qian2025toolrl, actless}.
Our work aims to enhance tool use capability by plan-guided data synthesis.

\section{Research Agent Learning}

\subsection{Problem Definition}\label{sec:problem_definition}
We characterize a Research Agent as a model that utilizes multi-turn tool calls to address user requests. 
We adopt the ReAct framework~\cite{react} to formalize the multi-turn setting: every turn is an 
iterative \thought-\action-\observation \ cycle. 
At each step, the model first produces an internal \thought, then executes an \action, and finally receives an \observation \  from the environment. 

Thus, a trajectory with $T$ turns can be denoted as
\[
y^T = (\tau_1, a_1, o_1, \ldots, \tau_T, a_T),
\]
where $\tau_i$ is the \thought, $a_i$ the \action, and $o_i$ the \observation\ at the $i$-th step. 
The process terminates at $a_T$ when the model produces the final answer.

To guide the model to follow this ReAct-style, we design a concise three-part prompt (detailed in Appendix~\ref{appendix:react_prompt}).
It consists of:
(1) \emph{Tool Descriptions}, which list the available tools and summarize their purpose and input parameters;
(2) \emph{Instructions}, which direct the model to follow the ReAct protocol and produce reasoning, actions, and observations step-by-step; and
(3) \emph{User Question}, which provides the actual query.
A minimal in-context example is also included to improve adherence to the format.

The action space consists of two external tools in addition to emitting the final answer: 
(i) \texttt{web\_search}, which queries a search engine and returns a list of top-$k$ candidate snippets with associated URLs, and 
(ii) \texttt{crawl\_webpage}, which extracts content from a chosen webpage. 

To make this behavior explicit in the generated trajectories, we wrap each component with special tags.
Specifically, reasoning tokens are enclosed in \think{,}, tool actions in \toolcall{,}, and tool responses in \toolresponse{,}.
When the agent decides to produce the final answer, the output is wrapped in \answer{,}.
This structured annotation helps the model internalize the ReAct reasoning pattern and ensures consistent tool use across training and inference.
Examples of the agent’s output following this format are provided in Appendix~\ref{appendix:case_study_1}.

% We represent each ReAct step by tagging its components with special tokens. The model's reasoning tokens is wrapped in \think{and}. 
% Tool-invoking actions are wrapped in \toolcall{\(\cdots\)} and encoded as a JSON object with fields 
% \texttt{name} (the tool identifier) and \texttt{parameters} (a key–value map of arguments); 
% the full schema and parameter constraints appear in Appendix~\ref{app:tool-spec}. 
% The corresponding tool output is wrapped as \toolresponse{f\(\cdots\)}. 
% When the agent decides to produce the final result, the answer is emitted as \answer{\(\cdots\)}.

% To elicit ReAct-style tool use, we employ a concise three-part prompt: \emph{Tool Descriptions}, \emph{Instructions}, and \emph{User Question}. 
% \emph{Tool Descriptions} list the available tools and summarize their purpose and required parameters. 
% \emph{Instructions} direct the LLM to follow the ReAct protocol and to format each component accordingly—for example, representing an action with the predefined \toolcall\{\} structure. 
% \emph{User Question} provides the query. 
% We also include a minimal in-context example to improve format adherence. 
% The complete prompt is given in Appendix~\ref{app:prompt}.

\begin{figure*}[t]
    \centering
    \includegraphics[width=0.9\linewidth]{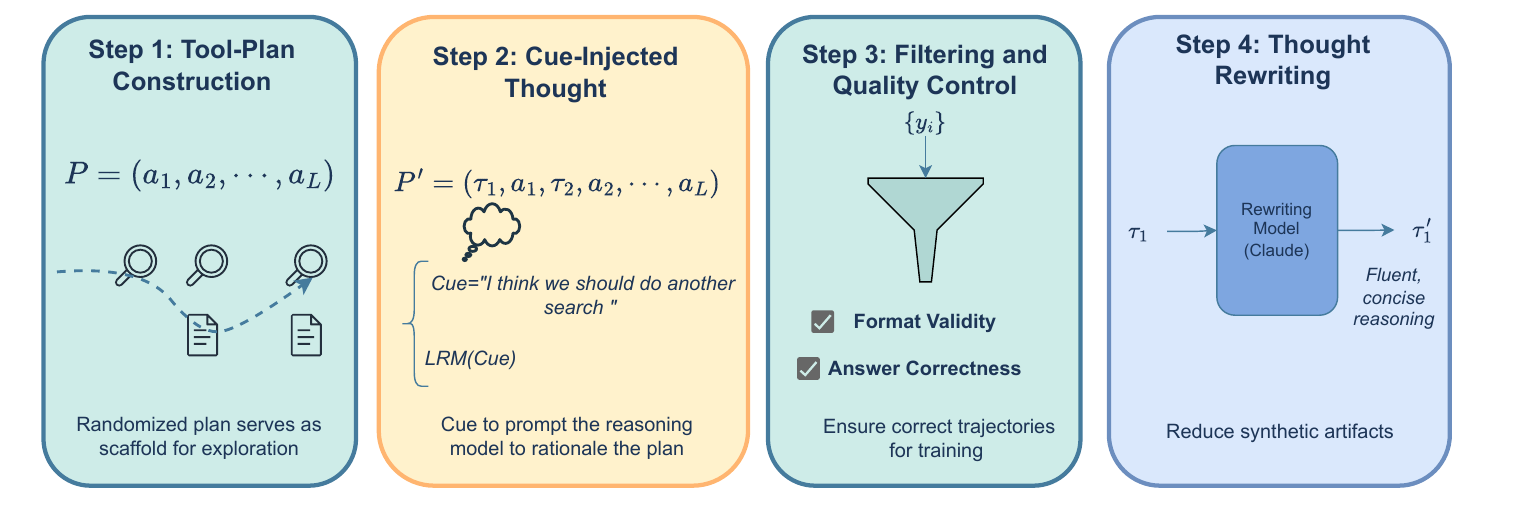}
    \caption{Overview of our data synthesis pipeline, which progresses from tool-plan generation, to cue-injected multi-step reasoning, and finally to trajectory filtering and quality control.}
    \label{fig:data-synthesis}
\end{figure*}

\subsection{Challenges in Agent Learning}
Research agents can be trained end-to-end with RLVR~\cite{lambert2024tulu3,kaufmann2025survey}.
When the model generates a trajectory following the ReAct structure with tags as specified in Sec.~\ref{sec:problem_definition}, we can parse the final answer from \answer{,} and compute the reward by exact match, F1 overlap or model judge based accuracy by comparing against the ground truth. 

% Given the prompting setup described in Sec.~\ref{sec:problem_definition}, the agent generates trajectories in the ReAct format and produces a final answer.
% This predicted answer is then compared with the ground-truth answer to obtain a scalar reward signal.
% Prior to RL, a cold-start SFT stage is typically applied to improve the success rate of tool usage and accelerate RL convergence~\cite{webthinker,simpledeepsearcher}.

However, RLVR training suffers from limitations inherent to its on-policy nature.
When the initial policy is poorly initialized or biased, the agent’s exploration becomes restricted, leading to stagnation around suboptimal behaviors and convergence to local optima.
% \zoey{Do we have data points for this: when we prompt the model or directly train it with RLVR the triggering rate of tool calls?}
We identify two prominent manifestations of this issue when training research agents:
(1) the agent tends to terminate reasoning prematurely, producing the final answer before executing enough tool calls to adequately explore the search space; and
(2) When multiple tools are available, the agent exhibits a strong preference for familiar ones—for instance, \texttt{web\_search} is invoked far more frequently than \texttt{crawl\_webpage}, which hampers deeper page understanding.
This shallow and imbalance in the empirical action distribution further reduces trajectory diversity and limits the effectiveness of subsequent RL updates.

\section{\framework~Framework}
%We adopt the RLVR paradigm under the ReAct framework to train our research agent. 
Our overall framework consists of two stages: a cold-start supervised fine-tuning (SFT) phase followed by reinforcement learning (RL) with outcome-based rewards. 
To improve the initial policy for deeper and explored tool use, we propose a novel \emph{plan-guided data generation} pipeline. 
This pipeline first prompts a large reasoning model (LRM) to generate multiple trajectories guided by randomized tool plans that encourage diverse tool-calling behaviors. 
We then apply a set of filtering rules to retain only trajectories with correct answers and a valid format for SFT training. 
% The details of this data generation process are presented in the next section. 
%After describing the data generation approach, we further introduce how the resulting data are used for cold-start SFT and how the agent is subsequently optimized via RL.

\subsection{Plan-Guided Data Synthesis}\label{sec:plan-guide-data-syn}
\paragraph{Step 1: Tool-Plan Construction.}  
We define a function \texttt{tool\_plan\_generator} that produces synthetic tool-use plans with simple but effective constraints. 
The length of the plan $L$ is first drawn from a uniform discrete range $[L_{\min}, L_{\max}]$. 
Each plan is then represented as a sequence $P = (a_1, \ldots, a_L)$, where the first action is always set to \texttt{web\_search} in order to bootstrap candidate evidence. 
For the remaining steps ($t \ge 2$), each action is sampled independently from the tool set $\{\texttt{web\_search}, \texttt{crawl\_webpage}\}$ with equal probability (0.5 in our case).  

These plans serve as scaffolds that encourage large reasoning models (LRMs) to produce trajectories with richer exploration behaviors, characterized by longer tool usage sequences and more diverse combinations of different tools.

\paragraph{Step 2: Cue-Injected Thoughts.}
The generated tool plan is injected into the initial user prompt as an auxiliary instruction.
However, this procedure faces two major challenges.
First, large reasoning models (LRMs) often fail to faithfully execute the tool-plan instructions, especially when the plan specifies multiple consecutive tool calls.
Second, the ReAct framework constrains the trajectory to a \thought-\action-\observation\ cycle, requiring the model to produce a natural-language thought before each tool call.
This structure makes it difficult to directly enforce the tool plan at the action level.

To address these issues, we design manually crafted \emph{cues} that are injected at the beginning of each thought step.
These cues serve as soft constraints, gently guiding the LRM toward the intended next action while preserving the natural ReAct reasoning flow.
For example, a cue may state “consider performing another search” or “examine one of the candidate webpages.”
Detailed cue templates are provided in Appendix~\ref{appendix:injected_cues}.

Formally, given a plan $P=(a_1,\ldots,a_L)$ with $a_t \in \{\texttt{web\_search},\texttt{crawl\_webpage}\}$, at step $t$ we first
create a \emph{thought prefix} inside the ReAct \singlethink\ block:
\[
c_t \;=\; \text{Cue}(a_t).
\]
We then let the LRMs \emph{continue} from this prefix to produce the remainder of the thought and, subsequently,
emit an action:
\[
(\Delta \tau_t,\; a_t)\; \sim\; \text{LRM} \!\left(\cdot \,\middle|\, H_{t-1},\; \texttt{<think> } c_t \right).
\]
The complete thought at step $t$ is given by $\tau_t = c_t \Vert \Delta \tau_t$,
where ``$\Vert$" denotes string concatenation.

% The realized thought is the model’s continuation of the cue,
% $\tau_t = c_t \,\Vert\, \Delta \tau_t$,
% followed by the structured \texttt{<tool\_call>} and its \texttt{<tool\_response>} per ReAct.
% The cue serves as a soft bias on the tool selection, while the subsequent continuation of the thought and the tool arguments (e.g., the formulation of the query and which URL to click) are still generated by the LRM.

\paragraph{Step 3: Filtering and Quality Control.}
For each question, we generate $N$ cue-injected ReAct trajectories $\{y_i\}_{i=1}^{N}$ from $N$ random tool plans.
%Let $\mathcal{Y}_q=\{y_i\}_{i=1}^{N}$ denote the trajectories for question $q$, where each $y_i$ contains a ReAct trajectory and a final answer.
We retain trajectories if they can pass the two checks:

\begin{itemize}
\item \textbf{Format validity.} The trajectory conforms to the ReAct schema (proper ordering and format of \singlethink, \singletoolcall, \singletoolresponse, \singleanswer).
\item \textbf{Answer correctness.} The final answer is verified as correct by the task-specific checker.
\end{itemize}

This filtering procedure ensures that the collected trajectories are both valid and diverse—since tool plans are sampled randomly, the retained set naturally covers a wide range of reasoning and exploration behaviors.

% We maintain multiple trajectories per question, provided they are well-formatted and yield the correct answer.
% Such diversity exposes the model, during cold-start SFT, to a broader range of exploration behaviors (e.g., longer search sequences, different search–click interleavings). We cap the number of accepted rollouts per question at a constant $M$ to prevent over-representation of any single pattern. 

\paragraph{Step 4: Thought Rewriting.}
Since the injected cues are long and manually constructed, they may introduce unnatural phrasing or stylistic bias. To mitigate this, we follow ~\cite{wu2025webdancerautonomousinformationseeking} to employ a high-quality rewriting model (Claude) to paraphrase each cue-injected thought into fluent, concise language while preserving its directive intent. This step reduces artifacts from synthetic construction and yields trajectories that are both structurally guided and linguistically natural, suitable for downstream SFT. The detailed of thought rewrite is provided in Appendix~\ref{appendix:thought-rewrite}.

\subsection{Reinforcement Learning with Cold-Start SFT}\label{method:rl-with-sft}
\paragraph{Cold-start SFT.} After constructing and filtering the synthetic trajectories described in Sec.~\ref{sec:plan-guide-data-syn}, we use them to perform cold-start SFT.
Starting from the question set $\mathcal{D}$, we collect the corresponding synthetic training data, resulting in a question-conditioned set $\mathcal{Y}^{+}_q=\{y_i\}$ for each $q\in\mathcal{D}$.
Because the synthesis uses zero-prompted LRMs, some questions admit no correct rollout; we remove those questions. The resulting supervision corpus is
$\mathcal{D}_{\text{sft}}=\{(q,y)\mid q\in\mathcal{D},\, y\in\mathcal{Y}^{+}_q\}$,
which may contain multiple valid trajectories per question.
We then train the policy $\pi_\theta$ using standard maximum-likelihood (cross-entropy) over this corpus.
The resulting model, denoted $\pi_{\text{sft}}$, serves as the initialization for the subsequent reinforcement learning stage.

% Importantly, prompting remains neutral at training time. During \emph{synthesis} we augment the base prompt with a tool plan (and cue-injected thoughts) solely to generate diverse/deeper trajectories; however, in both cold-start SFT and subsequent RL we revert to the \emph{base} prompt with exactly three components: (i) \emph{tool descriptions} (name, purpose, parameters), (ii) \emph{instructions} that prescribe the ReAct protocol and the required tag/JSON format, and (iii) the \emph{question} where the two prompting difference can be seen in Table~\ref{tab:prompt-two-col}. This separation avoids introducing any plan-specific bias: the agent learns from rich trajectories while being conditioned only on the standard prompt during supervision and later optimization.

\paragraph{RL.}
Let $\pi_{\text{sft}}$ denote the cold-start model obtained in the previous stage.
Starting from $\pi_{\text{sft}}$, we further optimize the policy by interacting with the tool-enabled environment.
For each prompt $x \in \mathcal{D}$—which includes the tool description, instruction and user question—we generate a \emph{group} of $G$ rollouts $\{y_i\}_{i=1}^{G}$ using the previous policy $\pi_{\theta_{\text{old}}}$.
Each rollout $y_i$ corresponds to a ReAct-style trajectory that involves tool usage and produces a final answer.

We optimize the policy using the following GRPO objective~\cite{shao2024deepseekmath}:
\begin{equation*}
\mathcal{J}_{\text{GRPO}}(\theta)
= \mathbb{E}_{x \sim \mathcal{D},\; \{y_i\}_{i=1}^G \sim \pi_{\theta_\text{old}}(\cdot|x)}
\left[ \hat{\mathcal{L}}_{\text{GRPO}}(\theta; x, y_i) \right].
\end{equation*}

\noindent where
\begin{equation*}
\begin{aligned}
\hat{\mathcal{L}}_{\text{GRPO}}(\theta)
&= \frac{1}{G} \sum_{i=1}^G
\frac{1}{\sum_{t=1}^{|y_i|} m_{i,t}}
\sum_{t=1}^{|y_i|} m_{i,t} \\
&\quad \cdot
\min\!\Big(
\rho_{i,t}(\theta) A_{i,t},
\mathrm{clip}(\rho_{i,t}(\theta), 1-\epsilon, 1+\epsilon) A_{i,t}
\Big).
\end{aligned}
\end{equation*}

Here, $A_{i,t}$ denotes the group-normalized relative advantage computed from the outcome reward of each rollout $y_i$. The importance ratio is defined as $\rho_{i,t}(\theta)=\frac{\pi_{\theta}(y_{i,t}\mid c_{i,t})}{\pi_{\theta_{\text{old}}}(y_{i,t}\mid c_{i,t})}$, and $\epsilon$ is the clipping hyperparameter from~\cite{schulman2017proximalpolicyoptimizationalgorithms}.
Following~\cite{search-r1}, we apply a binary mask $m_{i,t}\in \{0,1\}$, where $m_{i,t}=0$ if the token is obtained from tool responses and $m_{i,t}=1$ otherwise.

% Let $R_i=r(q, y_i)$ be the scalar reward, then the group-normalized relative advantage:
% \[
% A_i=\frac{R_i-\mathrm{mean}(\{R_j\}_{j=1}^{G})}{\mathrm{std}(\{R_j\}_{j=1}^{G})+\varepsilon}.
% \]
% We use the information mask~\cite{search-r1} 

% For each agent-authored token $y_{i,t}$ with context $c_{i,t}$ (history including any \texttt{<tool\_response>}), the importance ratio is
% \[
% \rho_{i,t}(\theta)=\frac{\pi_{\theta}(y_{i,t}\mid c_{i,t})}{\pi_{\theta_{\text{old}}}(y_{i,t}\mid c_{i,t})}.
% \]
% With a mask $m_{i,t}\in\{0,1\}$ that selects only agent tokens inside \texttt{<think>}, \texttt{<tool\_call>}, and \texttt{<answer>}, and a clipping parameter $\epsilon>0$, the GRPO objective for this task is

\emph{Reward shaping}
The outcome reward $r(q,y_i)$ is computed based on both answer accuracy and format validity.
We use the F1-score between the predicted and ground-truth answers, denoted as $s_{\text{ans}}$, to measure answer quality.
Let $f \in \{0,1\}$ indicate whether the model output follows the validated ReAct format.
The reward is defined as:
\[
r(q,y_i)=
\begin{cases}
s_{\text{ans}}+\alpha, & \text{if } f=1 \ \text{and}\ s_{\text{ans}}>0,\\[2pt]
0,                     & \text{if } f=1 \ \text{and}\ s_{\text{ans}}=0,\\[2pt]
0,                     & \text{if } f=0 \ \text{and}\ s_{\text{ans}}\ge \tau,\\[2pt]
-\alpha,               & \text{if } f=0 \ \text{and}\ s_{\text{ans}}< \tau.
\end{cases}
\]
Here, $\alpha$ is a format-related reward coefficient, and $\tau \in (0,1]$ is a threshold that determines what constitutes a “good” answer (set to $0.8$ in our experiments).

% This scheme (i) rewards correct answers that follow the ReAct format with a small format bonus, (ii) assigns zero reward to formatted but incorrect answers, (iii) prevents reward hacking by granting zero reward to unformatted but seemingly correct answers ($s_{\text{ans}}\ge\tau$), and (iv) penalizes unformatted failures.

\paragraph{Practical training tricks.}
% Multi-turn research agents often produce a non-trivial portion of \emph{void turns} (around $10\% \sim 15\%$ in our case)~\cite{xue2025simpletir}, where the trajectory fails to generate a final valid answer.
% Such failures typically occur when the trajectory exceeds the maximum token or turn limit, or when the agent produces an invalid tool call that does not conform to the required JSON schema.
% To enhance RL training stability and sample efficiency, we retain these trajectories for \emph{advantage} computation but exclude them from the policy loss.

% Concretely, let $\mathcal{U}$ denote the index set of trajectories with void turns within  a GRPO group.
% For each $i \in \mathcal{U}$, we set the loss mask to zero for all tokens, i.e., $\{m_{i,t}\}_{t=1}^{|y_i|}=0$, while still computing their outcome-based reward $r(q,y_i)$.
% Since these trajectories neither produce a valid answer nor follow the correct format, their reward is set to $r(q,y_i)=-\alpha$.

Multi-turn research agents often produce a non-trivial portion of \emph{void trajectories} (around $10\%\sim15\%$ in our case)~\cite{xue2025simpletir}, where the rollout fails to generate a valid final answer due to exceeding the maximum token or turn limit. To improve RL training stability and sample efficiency, we retain these trajectories for advantage computation while excluding them from the policy loss via loss masking.

In addition, trajectories that produce invalid tool calls—i.e., tool outputs that violate the required JSON schema—are treated as \emph{hard failures}. In such cases, we immediately terminate generation; the resulting rollouts do not follow the required format and are assigned a format-related penalty of $-\alpha$. We detail these two tricks in Appendix~\ref{appendix:trick-for-rl}.

% \zoey{Do we want to mention web data processing?}

% Requires:
% \usepackage{booktabs}
% \usepackage{tabularx}
% \usepackage{makecell}   % optional, for better line breaks in headers

% Requires:
% \usepackage{booktabs}
% \usepackage{tabularx}

% Requires:
% \usepackage{booktabs}
% \usepackage{tabularx}

\section{Main Results}
\subsection{Datasets}
\textbf{Evaluation} We evaluate on four multi-hop QA benchmarks—HotpotQA~\cite{hotpotqa}, 2WikiMultihopQA~\cite{2wiki}, MuSiQue~\cite{musique}, and Bamboogle~\cite{bamboogle}—and three advanced reasoning benchmarks that might need web search—GPQA~\cite{gpqa}, WebWalkerQA~\cite{webwalker}, and GAIA~\cite{gaia}. 
For each dataset, we use its official evaluation split (test if available, otherwise dev). 
To make evaluation more efficient, we downsample the evaluation sets of HotpotQA and 2WikiMultihopQA to 5K instances, while using the full evaluation splits for all other datasets.

\textbf{Training} For the SFT stage, we curate an 8k training set with the following composition: 70\% drawn from multi-hop QA sources (HotpotQA, 2WikiMultihopQA, MuSiQue, and MultiHop-RAG~\cite{multihop_rag}), 15\% from SuperGPQA~\cite{supergpqa}, and 15\% from WebWalker-Silver~\cite{webwalker}. For RL training, we construct another 8k set with the same proportions and augment it with 1.6k ARPO~\cite{arpo} examples, yielding approximately 9.6k ($\approx$10k) training instances in total. There is no overlap between the training sets and evaluation sets. 
% All sampling is performed at random without replacement, and evaluation splits are never used in training.

\subsection{Baselines}
We compare \framework\ against six baselines.
\begin{itemize}
    \item Direct Inference directly generates answers from the question without invoking any external tools. 
    \item Standard RAG~\cite{rag} performs single-turn retrieval-augmented generation, retrieving top-$k$ passages and conditioning the model’s response on them. 
    \item Search-o1~\cite{search-o1} is a multi-step search agent that decomposes a query into sub-queries, each handled by an auxiliary agent that performs interleaved reasoning and retrieval before returning summarized evidence to the main agent for subsequent reasoning steps. 
    \item Rejection Sampling~\cite{rft} uses the same synthetic data as \framework\ for cold-start but replaces RL with multi-iteration SFT with rejected sampling.
    \item Search-R1~\cite{search-r1} trains a multi-step search agent that interleaves reasoning and retrieval throughout the process, optimizing its final-answer accuracy via outcome-based reinforcement learning. 
    \item SimpleDeepSearcher~\cite{simpledeepsearcher} is an RL-based search agent that emphasizes building a high-quality synthetic dataset for robust cold-start initialization. 
\end{itemize}
% WebThinker~\cite{webthinker}, similar to Search-o1, employs an auxiliary agent for deep web exploration and summarization of retrieved results, while further improving performance by accumulating preference data from rollouts and applying DPO for policy refinement. 
% Finally,

\subsection{Implementation Details}

We use \textsc{Qwen3-32B} as the large reasoning model (LRM) for synthetic data generation. For each question, the tool-plan length $L$ is uniformly sampled from $[L_{\min}{=}3,\,L_{\max}{=}8]$, and a tool-dependent cue is injected at the beginning of every thought step. Each question produces $N{=}16$ rollouts, and we retain only those that satisfy both the correct ReAct format and the correct final answer. 

For cold-start SFT, we fine-tune the model for 2.5 epochs with a learning rate of $5e^{-6}$. For RL training, we adopt the GRPO algorithm with a rollout group size of $G{=}16$ and a learning rate of $1e^{-6}$. The maximum input length is 16K tokens, and the maximum number of tool-calling turns is 8. We use temperature~$=1.0$, top-$p{=}1.0$, and top-$k{=}{-}1$ in the rollout phase of RL training.
The reward combines format and correctness terms, with format weight $\alpha{=}0.2$ and correctness threshold $\tau{=}0.8$. 
% \hansi{more details of SFT + RL training. Such as verl's key parameters, how many documents return as the observation, Craw4AI's key parameters}

During inference, we follow the default thinking-mode decoding configuration recommended for Qwen3 models,\footnote{\url{https://huggingface.co/Qwen/Qwen3-8B}}
 using temperature~$=0.65$, top-$p{=}0.95$, and top-$k{=}20$. We keep the same token and turn limits as in training. Model performance is evaluated using the F1 score.

\subsection{Performance}
Table~\ref{tab:main-results} compares \framework with representative baselines across multi-hop and advanced QA benchmarks.
At the 8B scale, \framework achieves an average score of 0.580, clearly outperforming all competing methods.
Compared to rejection sampling with SFT (RFT), which attains 0.464, our approach improves by $25.0\%$.
This gain reflects RL’s higher sampling efficiency and its ability to learn from both positive and negative trajectories, unlike iterative SFT.
Relative to Search-R1, \framework gains $13.2\%$, showing that the cold-start SFT stage provides better policy initialization, which leads to more stable and efficient RL training.
Compared to SimpleDeepSearcher, which focuses on filtering high-quality questions, our approach gains $6.0\%$ by synthesizing more diverse trajectories during cold-start SFT.
We attribute this gain to the tool-plan-guided data synthesis, which exposes the model to a more diverse set of reasoning trajectories rather than a narrowly curated training corpus.
Similar trends hold at the 4B scale, where \framework continues to surpass RFT (+28.7\%), Search-R1 (+14.5\%), and SimpleDeepSearcher (+5.8\%) on average.

% The comparison between \framework and the baselines is presented in Table~\ref{tab:main-results}. Overall, \framework consistently outperforms all strong competitors across both multi-hop and advanced QA benchmarks. (1) Compared to Rejection Sampling with SFT (RFT), \framework achieves an average improvement of \textbf{xx\%} and \textbf{yy\%} for the 8B and 4B models, respectively, reflecting the superior sampling efficiency of our on-policy RL over iterative SFT. (2) Relative to Search-R1, \framework attains a substantial gain, underscoring the importance of cold-start SFT for establishing a stronger initial policy—an element absent in Search-R1. (3) \framework also consistently surpasses WebThinker across all benchmarks; since WebThinker optimizes its policy via DPO, an off-policy algorithm, it lacks the adaptive exploration benefits of our on-policy GRPO training. (4) Finally, compared to SimpleDeepSearcher, which also employs synthetic data for SFT, our method yields higher performance by explicitly planning tool usage during data generation, thereby encouraging deeper and more balanced tool-calling behaviors that better support RL refinement.
\begin{table*}[t]
\centering
\small
\setlength{\tabcolsep}{6pt}
\renewcommand\arraystretch{1.12}
\caption{Main results across benchmarks. Best performance in each column is \textbf{bold}.}
\label{tab:main-results}
\begin{tabular}{lcccccccc}
\toprule
\multirow{2}{*}{\textbf{Methods}} &
\multicolumn{4}{c}{\textbf{Multi-Hop QA}} &
\multicolumn{3}{c}{\textbf{Advanced QA}} & \\
\cmidrule(lr){2-5}\cmidrule(lr){6-8}\cmidrule(lr){8-8}
& \makecell{HotpotQA} & \makecell{2Wiki} & \makecell{Musique} & \makecell{Bamboogle} & \makecell{GPQA} & \makecell{WebWalkerQA} & \makecell{GAIA\\ Pass@4} & \makecell{\textbf{Avg.}} \\
\midrule
\multicolumn{9}{l}{\textbf{Qwen3\mbox{-}8B}} \\
% Qwen3-8B
Direct Inference     & .270 & .296 & .114 & .175 & .184 & .140 & .146 & .189 \\
RAG                  & .436 & .324 & .178 & .323 & .163 & .287 & .130 & .263 \\
Search-o1            & .424 & .517 & .216 & .546 & .324 & .339 & .257 & .375 \\
Rejection Sampling   & .543 & .625 & .234 & .652 & .425 & .353 & .417 & .464 \\
Search-R1            & .583 & .721 & .288 & .745 & .443 & .378 & .405 & .512 \\
SimpleDeepSearcher   & .608 & .779 & .323 & .765 & .466 & .424 & .452 & .547 \\
\rowcolor{lightblue}
\framework          & \textbf{.639} & \textbf{.821} & \textbf{.357} & \textbf{.784} & \textbf{.475} & \textbf{.474} & \textbf{.510} & \textbf{.580}
 \\
% \addlinespace[2pt]
\midrule
\multicolumn{9}{l}{\textbf{Qwen3\mbox{-}4B}} \\
% Qwen3-4B
Direct Inference     & .229 & .300 & .091 & .111 & .182 & .179 & .119 & .173 \\
RAG                  & .417 & .313 & .166 & .272 & .143 & .326 & .150 & .255 \\
Search-o1            & .417 & .503 & .204 & .453 & .239 & .323 & .250 & .341 \\
Rejection Sampling   & .525 & .588 & .221 & .637 & .249 & .345 & .312 & .411 \\
Search-R1            & .578 & .701 & .265 & .698 & .271 & .358 & .362 & .462 \\
SimpleDeepSearcher   & .589 & .742 & .298 & .738 & .312 & .403 & .415 & .500 \\
\rowcolor{lightblue}
\framework           & \textbf{.611} & \textbf{.793} & \textbf{.333} & \textbf{.753} & \textbf{.334} & \textbf{.438} & \textbf{.439} & \textbf{.529}
 \\

\bottomrule
\end{tabular}
\end{table*}

\section{Analysis}
\subsection{Tool Usage and Exploration}
We measure the relationship between the average number of tool calls (x-axis) and task performance (y-axis) across seven benchmarks, with the last panel reporting a macro-average. We compare our method, \framework, against two baselines: (i) RL from naive cold-start SFT and (ii) RL from scratch. For each method, we trace its trajectory by evaluating checkpoints at 20, 40, 60, and 80 RL steps, recording the average tool calls and the corresponding score at each checkpoint.

\begin{figure*}
    \centering
    \includegraphics[width=1.0\textwidth]{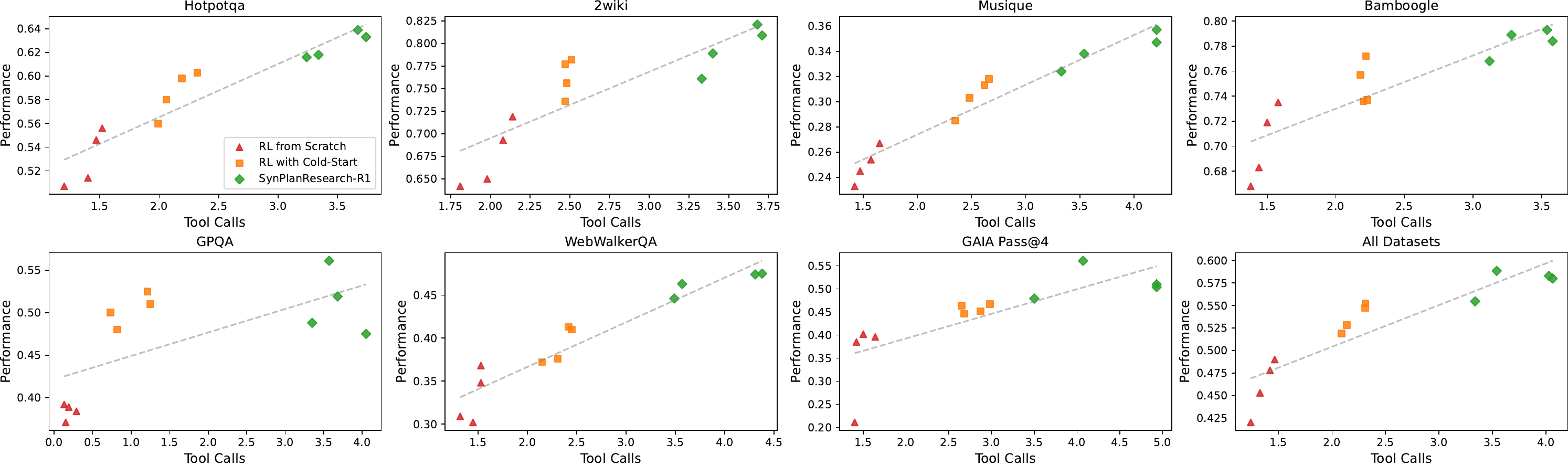}
    \caption{Relationship between average tool calls (x\mbox{-}axis) and task performance (y\mbox{-}axis; Exact Match, EM) across seven benchmarks; the last panel reports the macro average. Each method is shown at four RL checkpoints (20, 40, 60, and 80 steps).}

    \label{fig:tool_vs_perf}
\end{figure*}

Figure~\ref{fig:tool_vs_perf} reveals a clear relationship between exploration depth and performance: methods that make more tool calls consistently achieve higher accuracy. While all methods increase their tool usage as training progresses, RL-from-scratch and RL from naive cold-start SFT remain limited in both early and final exploration depth, resulting in inferior performance compared to \framework. This suggests that RL alone can refine an existing exploration policy but is insufficient to overcome a shallow tool-use prior induced by initialization; in contrast, \framework provides a stronger exploration prior that enables substantially deeper search and a higher performance ceiling.

Moreover, the exploration budget adapts to dataset difficulty. Harder web-based tasks like GAIA require more iterative evidence gathering, whereas curated multi-hop QA tasks saturate with fewer tool calls. This indicates that \framework allows the agent to dynamically modulate its search depth based on task complexity.

% In summary, Figure~\ref{fig:tool_vs_perf} shows that \framework\ addresses the limitation of shallow tool use: by seeding the agent with stronger exploration priors at cold start, it enables more effective RL fine-tuning, leading to deeper, more adaptive tool usage and superior performance across benchmarks.

% ---------- in body ----------

\begin{figure}
    \centering
    \includegraphics[width=0.5\textwidth]{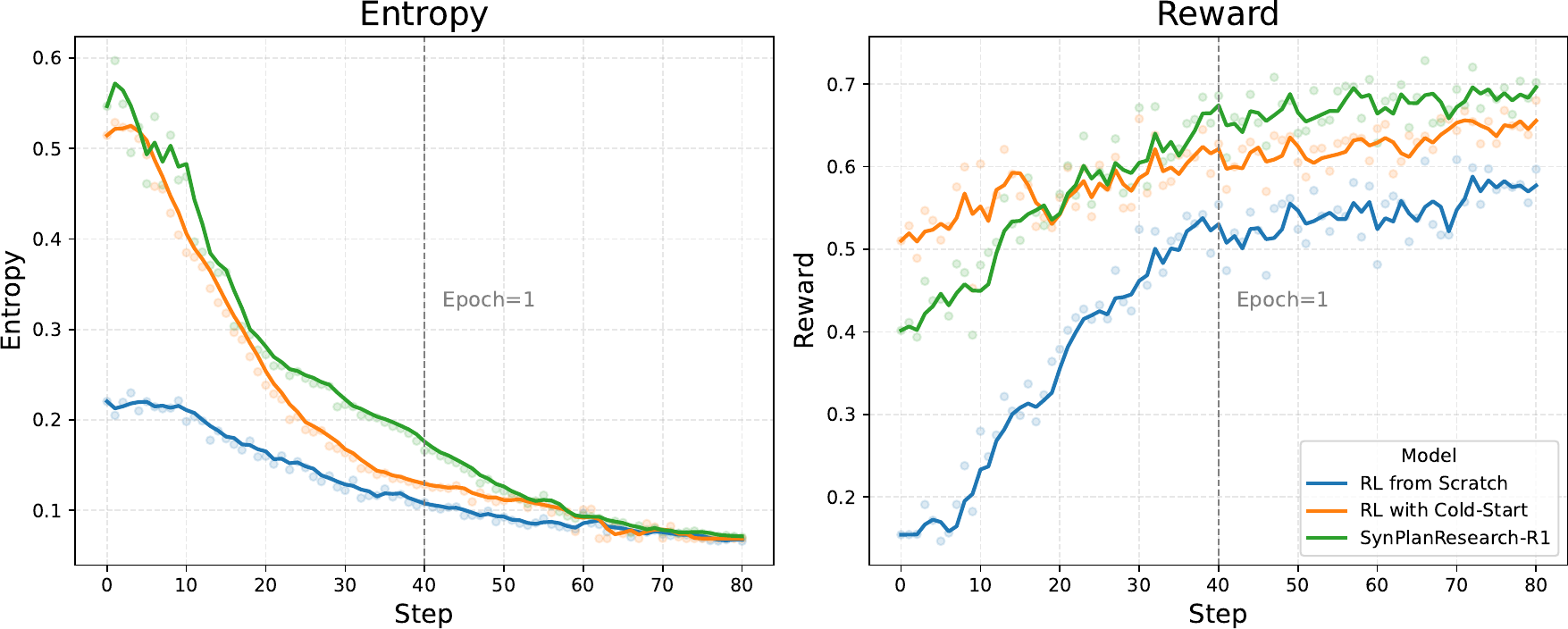}
    \caption{Training dynamics of policy entropy (left) and reward (right) over RL steps for three regimes.}

    \label{fig:train-dynamic}
\end{figure}
\subsection{Training Dynamics: Entropy and Reward}
To further understand the learning behavior of different methods, we analyze their training dynamics by tracking key metrics throughout the RL optimization process. Figure~\ref{fig:train-dynamic} illustrates the evolution of policy entropy (left) and training reward (right) across three approaches: our proposed \framework, RL from naive cold-start SFT, and RL from scratch.

The entropy metric reveals crucial differences in exploration capability. We observe that \framework maintains consistently higher entropy throughout training compared to both baselines. This elevated entropy indicates a more diverse and exploratory policy, directly resulting from our synthetic plan–guided initialization. The strong exploration priors embedded during cold-start SFT enable the agent to maintain a broader distribution over possible actions, rather than prematurely converging to limited tool-use patterns.

The reward dynamics (Figure~\ref{fig:train-dynamic}, right) reveal an interesting learning trajectory. While \framework initially lags behind the naive SFT+RL baseline in early training stages, it demonstrates stronger long-term improvement and eventually surpasses all other methods. This pattern aligns with the entropy observations: the increased exploration capacity initially leads to more stochastic behavior and slower reward accumulation, but ultimately enables the discovery of superior strategies that yield higher final performance. In contrast, the naïve SFT+RL approach plateaus earlier, suggesting limited capacity to escape local optima due to insufficient exploration diversity.

% These findings collectively demonstrate that our method’s enhanced exploration capability, fostered through carefully designed initialization, enables more effective policy optimization. Though initially slower to capitalize on rewards, the sustained exploration allows \framework to discover more robust and higher-performing strategies over extended training.

\subsection{Ablation Study}
\begin{table}[t]
\centering
\footnotesize
\setlength{\tabcolsep}{3.8pt}
\renewcommand\arraystretch{1.12}
\caption{Ablation of \framework\ across benchmarks. 
Best value in each column is \textbf{bold}.}
\label{tab:ablations-synplan}

\begin{tabularx}{\linewidth}{>{\scriptsize}l *{4}{S}} % �� 第一列单独缩小字体
\toprule
\textbf{Method} & {MulHop} & {GPQA} & {WebWlk.} & {GAIA@4} \\
\midrule
\rowcolor{lightblue}
\textbf{\framework} & \bfseries 0.650 & 0.475 & \bfseries 0.474 & \bfseries 0.510 \\

\addlinespace[2pt]
\rowcolor{gray!10}
\multicolumn{5}{l}{\emph{RL Training}} \\
\scriptsize \quad 1.\;max\_turns (8$\rightarrow$6)   & 0.624 & 0.438 & 0.456 & 0.502 \\
\scriptsize \quad 2.\;-- explore\_rollout           & 0.592 & 0.414 & 0.375 & 0.455 \\
\scriptsize \quad 3.\;-- filter\_unfinished         & 0.631 & 0.457 & 0.432 & 0.497 \\

\addlinespace[2pt]
\rowcolor{gray!10}
\multicolumn{5}{l}{\emph{Cold-start SFT}} \\
\scriptsize \quad 4.\;-- claude\_rewrite             & 0.648 & 0.471 & 0.461 & 0.507 \\
\scriptsize \quad 5.\;-- cue\_injected thoughts      & 0.609 & \bfseries 0.511 & 0.411 & 0.472 \\

\addlinespace[2pt]
\rowcolor{gray!10}
\multicolumn{5}{l}{\emph{Used Tool}} \\
\scriptsize \quad 6.\;only search                   & 0.626 & 0.374 & 0.442 & 0.430 \\
\bottomrule
\end{tabularx}
\end{table}

\textbf{RL training choices.}
\emph{Row 1} reduces the maximum tool-use budget from 8 to 6 steps and consistently degrades performance (e.g., $-0.26$ on Multi-Hop QA, $-0.18$ on WebWalkerQA), indicating a positive correlation between performance and the depth of tool-based exploration.
\emph{Row 2} lowers the rollout-time temperature from 1.0 to 0.6, resulting in the largest performance drops across tasks (e.g. $-0.65$ on Multi-Hop QA, $-0.99$ on WebWalkerQA). This highlights the importance of maintaining sufficient exploration diversity during RL.
\emph{Row 3} treats unfinished rollouts (exceeding token or turn limits) as negative examples instead of filtering them out, also leading to performance declines (e.g., $-0.13$ on GAIA) and reflecting reduced training stability.

\textbf{Cold-start SFT components.}
\emph{Row 4} removes Claude-based thought rewriting, which shortens verbose reasoning traces into more compact and natural prefixes. Its impact on accuracy is minor (e.g., $-0.02$ on Multi-Hop QA, $-0.03$ on WebWalkerQA), suggesting that performance is driven primarily by whether reasoning induces correct tool usage rather than by linguistic concision.
In contrast, \emph{Row 5} removes cue-injected thoughts during synthetic trajectory generation, eliminating explicit step-wise guidance that anchors the model to a target tool sequence. This ablation causes substantial performance drops on three of four benchmarks (Multi-Hop QA: $-0.31$, WebWalkerQA: $-0.63$, GAIA: $-0.38$), indicating that cues play a central role in shaping effective exploration behavior during cold-start SFT. Without cues, the model tends to deviate from the prescribed diverse tool plan, resulting in shallower exploration patterns that persist into RL.

\textbf{Tool use.}
\emph{Row 6} restricts the agent to the web search tool only. This leads to a pronounced degradation on GAIA ($-0.80$), demonstrating that webpage crawling is a critical capability for complex web-based reasoning tasks that require deeper evidence gathering beyond search snippets.

\subsection{Pre-training Statistics}
We next examine how different zero-shot prompting schemes shape the behavior of the LRM that we use to synthesize trajectories. As shown in Table~\ref{tab:pretrain-stats}, the vanilla ReAct prompt yields short and weak exploratory traces: the model issues only 1.35 tool calls on average and seldomly performs crawling actions (0.05 calls per query). Adding an explicit tool plan nudges the LRM toward richer exploration: it substantially increases web crawling actions and raises the total number of tool calls from 1.400 to 2.050, but the generated trajectories follow the prescribed plan only 28.32\% of the time, indicating that plans alone are not sufficient for reliable execution. When we further inject step-wise cues, the LRM not only increases its overall tool usage (4.36 calls per query with substantially more searches and crawling), but also adheres to the target plan in 76.96\% of trajectories. This confirms our motivation that cue-injected plans can imprint a much stronger exploration prior into the synthetic data, which later manifests as higher entropy and deeper tool usage during RL training, as observed in our analysis of tool-usage and performance curves.

\begin{table}[t]
\centering
\footnotesize
\setlength{\tabcolsep}{3.8pt}
\renewcommand\arraystretch{1.12}
\caption{Average numbers of tool calls, Web Search and Web Crawl calls, and  produced under different zero-shot prompting methods. Acc measures how well the generated trajectories follow the tool plan specified in the prompt.}
\label{tab:pretrain-stats}

\begin{tabularx}{\linewidth}{>{\small}l *{3}{S} >{\small}r}
\toprule
\textbf{Method} 
& {\# Search} 
& {\# Crawl} 
& {\# Total} 
& Accuracy \\
\midrule
Standard ReAct 
& 1.35 & 0.05 & 1.40 & -- \\

+ Tool Plan 
& 1.48 & 0.57 & 2.05 & 28.32\% \\

+ Tool Plan + Cue
& 2.72 & 1.64 & 4.36 & 76.96\% \\
\bottomrule
\end{tabularx}
\end{table}

\section{Conclusion}
We propose \framework, a plan-guided synthetic data framework that improves web research agents by shaping exploration behavior during the cold-start SFT phase.
By controlling downstream RL and varying only the SFT initialization, we show that \framework induces deeper tool-use exploration, avoiding the shallow and premature termination patterns common under naive initialization, and leads to consistently stronger final performance.
Across seven challenging open-domain QA and web-based benchmarks, \framework significantly outperforms state-of-the-art baselines, highlighting cold-start SFT as a critical lever for training robust web research agents.

% Acknowledgements should only appear in the accepted version.
% \section*{Acknowledgements}

% \textbf{Do not} include acknowledgements in the initial version of
% the paper submitted for blind review.

% If a paper is accepted, the final camera-ready version can (and
% usually should) include acknowledgements.  Such acknowledgements
% should be placed at the end of the section, in an unnumbered section
% that does not count towards the paper page limit. Typically, this will 
% include thanks to reviewers who gave useful comments, to colleagues 
% who contributed to the ideas, and to funding agencies and corporate 
% sponsors that provided financial support.

\section*{Impact Statement}

This paper presents a methodological contribution aimed at improving the training of research agents with multi-turn tool use. 
The techniques developed in this work are intended to support general-purpose research agents for tasks such as information seeking and question answering. While such systems may be applied in a wide range of downstream applications, the methods themselves do not introduce new forms of data collection, user profiling, or decision-making beyond existing tool-based language model systems. As such, we do not foresee novel ethical risks specific to our approach beyond those already associated with large language models and web-based information retrieval.
Overall, we believe this work contributes positively to the development of more robust and controllable machine learning systems, and we do not identify any broader societal impacts that require specific discussion beyond established considerations in the field.

% In the unusual situation where you want a paper to appear in the
% references without citing it in the main text, use \nocite
% \nocite{langley00}

\bibliography{example_paper}
\bibliographystyle{icml2025}

%%%%%%%%%%%%%%%%%%%%%%%%%%%%%%%%%%%%%%%%%%%%%%%%%%%%%%%%%%%%%%%%%%%%%%%%%%%%%%%
%%%%%%%%%%%%%%%%%%%%%%%%%%%%%%%%%%%%%%%%%%%%%%%%%%%%%%%%%%%%%%%%%%%%%%%%%%%%%%%
% APPENDIX
%%%%%%%%%%%%%%%%%%%%%%%%%%%%%%%%%%%%%%%%%%%%%%%%%%%%%%%%%%%%%%%%%%%%%%%%%%%%%%%
%%%%%%%%%%%%%%%%%%%%%%%%%%%%%%%%%%%%%%%%%%%%%%%%%%%%%%%%%%%%%%%%%%%%%%%%%%%%%%%
\newpage
\appendix
\onecolumn
\newpage
\appendix
\onecolumn
\section{Prompts}
\subsection{ReAct Prompt Template for Research Agent}~\label{appendix:react_prompt}

\begin{tcolorbox}[
  enhanced,
  breakable,
  colback=black!3,
  colframe=black!25,
  title={ReAct Prompt Template for Research Agent},
  colbacktitle=black!70,
  coltitle=white,
  fonttitle=\bfseries,
  title filled,             % 让标题条是实心色块
  minipage boxed title,     % 标题宽度=整盒宽度
  arc=6pt,
  % drop fuzzy shadow=black!30,
  left=8pt,right=8pt,top=6pt,bottom=8pt,
  width=\textwidth
]
\small  % 调小字体，使宽行不超出
\setlength{\parskip}{3pt}

You are an intelligent agent that can interact with tools to answer complex questions. Below are the available tools: \\[6pt]
\texttt{<tool>} \\
Tool Name: web\_search \\
Description: Search for information from the internet. \\
Usage: This tool lets the agent interact with the \texttt{web\_search} API. \\
Parameters:
\texttt{\{"type":"object","properties":{"query":{"type":"string"}},"required":["query"]\}} \\
\texttt{</tool>} \\
\texttt{<tool>} \\
Tool Name: crawl\_webpage \\
Description: A tool for fetching content of a webpage based on its URL. \\
Usage: This tool lets the agent interact with the \texttt{crawl\_webpage} API. \\
Parameters:
\texttt{\{"type":"object","properties":{"url":{"type":"string","description":"The url of a webpage to fetch"}},"required":["url"]\}} \\
\texttt{</tool>} \\[6pt]
Instructions: \\
You starts with one or more cycles of (thinking about which tool to use → performing tool call → waiting for tool response), and ends with (thinking about the answer → answer of the question). 
The thinking processes, tool calls, tool responses, and answer are enclosed within their tags. There could be multiple thinking processes, tool calls, tool call parameters and tool response parameters.
The tool you can should use is one of the following: web\_search, crawl\_webpage. \\
When you reach the \texttt{<answer>} tag, your final answer MUST be short and direct — ideally one phrase or one sentence. DO NOT repeat the question or explain your reasoning again. The reasoning belongs only in \texttt{<think>}.
For example, when question is who sings Beat it, your answer should be \texttt{<answer>} Michael Jackson \texttt{</answer>}. \\[6pt]

Example response:

\noindent
\texttt{<think>} thinking process here \texttt{</think>}\\
\texttt{<tool\_call>}\\
\texttt{\{}%
\texttt{\{}"name": "tool name here", "parameters": \texttt{\{}"parameter name here": parameter value here, "another parameter name here": another parameter value here, ...\texttt{\}}\texttt{\}}\texttt{\}}\\
\texttt{</tool\_call>}\\
\texttt{<tool\_response>}\\
tool\_response here\\
\texttt{</tool\_response>}\\
\texttt{<think>} thinking process here \texttt{</think>}\\
\texttt{<tool\_call>}\\
\texttt{\{}%
\texttt{\{}"name": "another tool name here", "arguments": \texttt{\{}...\texttt{\}}\texttt{\}}\texttt{\}}\\
\texttt{</tool\_call>}\\
\texttt{<tool\_response>}\\
tool\_response here\\
\texttt{</tool\_response>}\\
(more thinking processes, tool calls and tool responses here)\\
\texttt{<think>} thinking process here \texttt{</think>}\\
\texttt{<answer>} answer here \texttt{</answer>}\\[6pt]

Question: \texttt{\{\}}

\end{tcolorbox}

\captionof{table}{full React-style prompt used to train and evaluate \framework. 
It includes (1) tool descriptions, (2) ReAct protocol instructions, and (3) the user question.}
\label{tab:react_prompt}

\subsection{ReAct Prompt Template with a Inserted Tool Plan for LRMs}~\label{appendix:react_prompt_with_tool_plan}

\begin{tcolorbox}[
  enhanced,
  breakable,
  colback=black!3,
  colframe=black!25,
  title={ReAct Prompt Template with a Inserted Tool Plan for LRMs},
  colbacktitle=black!70,
  coltitle=white,
  fonttitle=\bfseries,
  title filled,             % 让标题条是实心色块
  minipage boxed title,     % 标题宽度=整盒宽度
  arc=6pt,
  % drop fuzzy shadow=black!30,
  left=8pt,right=8pt,top=6pt,bottom=8pt,
  width=\textwidth
]
\small  % 调小字体，使宽行不超出
\setlength{\parskip}{3pt}

You are an intelligent agent that can interact with tools to answer complex questions. Below are the available tools: \\[6pt]
\texttt{<tool>} \\
Tool Name: web\_search \\
Description: Search for information from the internet. \\
Usage: This tool lets the agent interact with the \texttt{web\_search} API. \\
Parameters:
\texttt{\{"type":"object","properties":{"query":{"type":"string"}},"required":["query"]\}} \\
\texttt{</tool>} \\
\texttt{<tool>} \\
Tool Name: crawl\_webpage \\
Description: A tool for fetching content of a webpage based on its URL. \\
Usage: This tool lets the agent interact with the \texttt{crawl\_webpage} API. \\
Parameters:
\texttt{\{"type":"object","properties":{"url":{"type":"string","description":"The url of a webpage to fetch"}},"required":["url"]\}} \\
\texttt{</tool>} \\[6pt]
Instructions: \\
You start with one or more cycles of (thinking about which tool to use → performing tool call → waiting for tool response), and end with (thinking about the answer → answer of the question). 
The thinking processes, tool calls, tool responses, and answer are enclosed within their tags. There could be multiple thinking processes, tool calls, tool call parameters and tool response parameters. \\
The tool you can use should be one of the following: \texttt{web\_search}, \texttt{crawl\_webpage}. \\[3pt]

Below is a tool usage plan you should follow (in order): \texttt{web\_search → crawl\_webpage → crawl\_webpage → web\_search} \\[6pt]

When you reach the \texttt{<answer>} tag, your final answer MUST be short and direct — ideally one phrase or one sentence. DO NOT repeat the question or explain your reasoning again. The reasoning belongs only in \texttt{<think>}.
For example, when question is who sings Beat it, your answer should be \texttt{<answer>} Michael Jackson \texttt{</answer>}. \\[6pt]

Example response:

\noindent
\texttt{<think>} thinking process here \texttt{</think>}\\
\texttt{<tool\_call>}\\
\texttt{\{}%
\texttt{\{}"name": "tool name here", "parameters": \texttt{\{}"parameter name here": parameter value here, "another parameter name here": another parameter value here, ...\texttt{\}}\texttt{\}}\texttt{\}}\\
\texttt{</tool\_call>}\\
\texttt{<tool\_response>}\\
tool\_response here\\
\texttt{</tool\_response>}\\
\texttt{<think>} thinking process here \texttt{</think>}\\
\texttt{<tool\_call>}\\
\texttt{\{}%
\texttt{\{}"name": "another tool name here", "arguments": \texttt{\{}...\texttt{\}}\texttt{\}}\texttt{\}}\\
\texttt{</tool\_call>}\\
\texttt{<tool\_response>}\\
tool\_response here\\
\texttt{</tool\_response>}\\
(more thinking processes, tool calls and tool responses here)\\
\texttt{<think>} thinking process here \texttt{</think>}\\
\texttt{<answer>} answer here \texttt{</answer>}\\[6pt]

Question: \texttt{\{\}}

\end{tcolorbox}

\captionof{table}{A full ReAct prompt template with an instantiated tool plan used to guide LRMs toward generating diverse trajectories.
We initialize an explicit tool plan as: web\_search → crawl\_webpage → crawl\_webpage → web\_search. Note that the tool plan is randomly generated by for each example.}
\label{tab:react_prompt_with_tool_calls}

\subsection{Injected Cues for Each Action}~\label{appendix:injected_cues}

To stabilize plan-conditioned reasoning and encourage the LRM to faithfully follow the prescribed tool plan, we inject short cue sentences at the beginning of each ReAct thought step. These cues act as soft constraints: they gently nudge the model toward the intended next action while preserving the natural language flow of ReAct-style reasoning.
Table~\ref{tab:inject_cues} lists the exact cue templates we use for each tool action.

\begin{table}[h!]~\label{table:injected_cues}
\centering
\small
\caption{Injected cue templates used at the beginning of each ReAct \texttt{<think>} step.}
\begin{tabular}{p{0.18\linewidth} p{0.75\linewidth}}
\toprule
\textbf{Action} & \textbf{Injected Cue Template} \\
\midrule
\texttt{web\_search} & 
To answer this, I probably need to gather external information first. 
Let me think about the best way to phrase the query for web\_search. \\[6pt]

\texttt{crawl\_webpage} & 
There are several promising links from the search results. 
Perhaps I should inspect one of them to obtain more specific information. \\
\bottomrule
\end{tabular}

\label{tab:inject_cues}
\end{table}

\subsection{Thought Rewrite}
\label{appendix:thought-rewrite}

Given a ReAct-style trajectory 
\(H_{L} = \{(\tau_t, a_t, o_t)\}_{t=1}^{L}\),
where \(\tau_t\), \(a_t\), and \(o_t\) denote the thought, action, and observation at step \(t\), we rewrite the thoughts using the Claude-3.7-Sonnet\footnote{us.anthropic.claude-3-7-sonnet-20250219-v1:0}.
For each step \(t\), we generate a rewritten thought \(\tilde \tau_t\) that (i) remains consistent with the original action \(a_t\) and observation \(o_t\), and (ii) is concise and natural.
We process the trajectory sequentially.
At step \(t\), the rewrite model is given:

\begin{itemize}
    \item the user prompt \(q\),
    \item the history of previously rewritten steps \(\{(\tilde \tau_i, a_i, o_i)\}_{i < t}\), and
    \item the current action \(a_t\) (and the latest observation \(o_{t-1}\)),
\end{itemize}

and is asked to output only the next rewritten thought \(\tilde \tau_t\) that explains why taking action \(a_t\) is reasonable at this point.
Formally, the rewriting model \(\mathrm{LRM}_{\mathrm{rewrite}}\) is queried as
\[
  \tilde \tau_t 
  \;=\; \mathrm{LRM}_{\mathrm{rewrite}}\bigl(q,\{(\tilde \tau_i,a_i,o_i)\}_{i < t}, a_t \bigr),
\]
and we replace the original thought \(\tau_t\) in the trajectory with \(\tilde \tau_t\).
We repeat this procedure for all steps \(t = 1,\dots,L\).

The concrete prompt used for the teacher LLM at step \(t\) is in Table~\ref{tab:thought-rewrite}.

\begin{tcolorbox}[
  enhanced,
  breakable,
  colback=black!3,
  colframe=black!25,
  title={Thought Rewrite Prompt for Step $t$},
  colbacktitle=black!70,
  coltitle=white,
  fonttitle=\bfseries,
  title filled,
  minipage boxed title,
  arc=6pt,
  left=8pt,right=8pt,top=6pt,bottom=8pt,
  width=\textwidth
]
\small
\setlength{\parskip}{3pt}

You are simulating the internal reasoning process of a strong LLM agent.

You will be given the reasoning trajectory of a \emph{weaker open-source LLM agent} that attempts to complete an information-seeking task in multiple steps.
This agent interacts with external tools in a multi-turn format:
\texttt{Thought $\rightarrow$ Action $\rightarrow$ Observation}, aiming to answer a user's question.

Each step is formatted as: \\
\texttt{Thought: ...} \\
\texttt{Action: ...} \\
\texttt{Observation: ...}

---

Your task is to generate a rewritten version of the current \textbf{Thought}, called \texttt{Your\_Rewrite\_Thought}, which explains the rationale behind the \textbf{current Action} in a concise and natural way.

You will be provided with:
\begin{itemize}
  \item the \textbf{user prompt} that defines the overall goal,
  \item the previous history consisting of \texttt{Your\_Rewrite\_Thought}, \texttt{Action}, and \texttt{Observation} steps,
  \item the \textbf{current Action} taken by the agent, and
  \item the latest \textbf{Observation} from the previous step.
\end{itemize}

\textbf{Instructions:}
\begin{enumerate}
  \item \texttt{Your\_Rewrite\_Thought} should simulate how an LLM would think \emph{before} taking the current Action.
  \item The thought must logically justify the current Action, without changing or contradicting it.
  \item Keep it concise and natural; avoid overly verbose, stylistic, or redundant explanations.
  \item Do \emph{not} rewrite earlier thoughts.
  \item Output \emph{only} the rewritten thought text --- do not include any labels or prefixes such as
        ``\texttt{Your\_Rewrite\_Thought:}'', ``\texttt{Thought:}'', ``\texttt{Action:}'', or ``\texttt{Observation:}''.
  \item The literal string ``\texttt{Your\_Rewrite\_Thought}'' must \emph{not} appear anywhere in your output.
\end{enumerate}

---

User Prompt: \texttt{"<user\_prompt>"} \\[4pt]

[Begin Trajectory So Far] \\
Step $0$: \\
\quad \texttt{Your\_Rewrite\_Thought:} $\tilde h_0$ \\
\quad \texttt{Action:} $a_0$ \\
\quad \texttt{Observation:} $o_0$ \\
\vdots \\
Step $t-1$: \\
\quad \texttt{Your\_Rewrite\_Thought:} $\tilde h_{t-1}$ \\
\quad \texttt{Action:} $a_{t-1}$ \\
\quad \texttt{Observation:} $o_{t-1}$ \\
\mbox{[End Trajectory]}\\[4pt]

Step $t$: \\
\quad \texttt{Action:} $a_t$ \\[2pt]
\texttt{Your\_Rewrite\_Thought:}
\end{tcolorbox}
\captionof{table}{Thought rewrite prompt for step $t$.}
\label{tab:thought-rewrite}

\section{Data Synthesis Algorithm}

\begin{algorithm}[t]
\caption{\textbf{SynPlanResearch-R1}: Plan-Guided Synthetic ReAct Data Synthesis}
\label{alg:synplan_synthesize_dataset}
\begin{algorithmic}[1]
\REQUIRE
Dataset $\mathcal{D}$, where each element is a pair $(q, \texttt{gold\_ans})$ consisting of
a question $q$ and its corresponding golden answers \texttt{gold\_ans};
rollout budget $N$;
plan length range $[L_{\min}, L_{\max}]$.

\ENSURE
Cold-start SFT corpus $\mathcal{D}_{\text{sft}}$.

\vspace{0.5em}
\STATE Initialize $\mathcal{D}_{\text{sft}} \leftarrow \emptyset$.
\FOR{each ( $q, \texttt{gold\_ans})\in \mathcal{D}$}
    \STATE Initialize $\mathcal{S}(q) \leftarrow \emptyset$ \algcomment{candidate trajectories for $q$}

    \FOR{$i=1$ to $N$}
        \STATE \algcomment{Sec.~4.1, Step~1: tool plan consturction}
        \STATE Sample $L \sim \mathrm{Uniform}\{L_{\min},\ldots,L_{\max}\}$.

        \STATE Sample a tool plan $P=(a_1,\ldots,a_L)$ with $a_1=\texttt{web\_search}$ and
               $a_t \sim \mathcal{A}$ for $t\ge 2$.

        \STATE Construct the initial ReAct prompt by injecting $P$ into the plan-conditioned
               template and appending query $q$; denote it as $H_0$ (Appendix~\ref{appendix:react_prompt_with_tool_plan}).
        \STATE \algcomment{Sec.~4.1, Step~2: Cue-Injected Thougths}

        \FOR{$t=1$ to $L$}
            \STATE $c_t \leftarrow \mathrm{Cue}(a_t)$ using the action-specific cue template (Appendix~\ref{appendix:injected_cues}).

            \STATE Query the LRM to complete the current ReAct step from the prefix inside \texttt{<think>}:
            \[
            (\Delta\tau_t,\ a_t)\sim
            \mathrm{LRM}\big(\cdot \mid H_{t-1}, \langle\texttt{think}\rangle c_t\big),
            \]
            where $\Delta\tau_t$ is the remainder of the thought text and $a_t$ is the action (tool call)

            \STATE Execute $a_t$ to obtain observation $o_t$ (tool response),

            \STATE Form the full thought block
                   $\tau_t \leftarrow \langle\texttt{think}\rangle c_t \,\|\, \Delta\tau_t$.

            \STATE Update history
                   $H_t \leftarrow H_{t-1} \oplus (\tau_t,\ a_t,\ o_t)$.
        \ENDFOR

        \STATE Continue the LRM for one final step (no tool call) to produce the terminal
               \texttt{<think>} and \texttt{<answer>} blocks:
        \[
        (\tau_{L+1},\ \texttt{ans}) \sim \mathrm{LRM}(\cdot \mid H_L),
        \]
        and set $H_{L+1} \leftarrow H_L \oplus (\tau_{L+1},\ \texttt{ans})$.

        \STATE Parse $\texttt{ans}$ from the \texttt{<answer>} tag.

        \STATE \algcomment{Section 4.1, Step 3: Filtering and Quality Control}
        \STATE Run \textsc{FormatCheck}$(H_{L+1})$ and
               \textsc{AnswerCheck}$(\texttt{ans},\ \texttt{gold\_ans})$

        \IF{\textsc{FormatCheck} fails \OR \textsc{AnswerCheck} fails}
            \STATE \textbf{continue}
        \ENDIF

        \STATE \algcomment{Section 4.1, Step 4: Thougth Rewrite}
        \STATE Rewrite each thought step sequentially (Appendix~\ref{appendix:thought-rewrite}):
        \[
        \tilde{H}_{L+1} \leftarrow \textsc{RewriteThoughts}(q,\ H_{L+1}),
        \]
        where only thought text is rewritten while keeping all actions,
        observations, and the final answer unchanged.

        \STATE \STATE $\mathcal{D}_{\text{sft}} \leftarrow \mathcal{D}_{\text{sft}} \cup \tilde{H}_{L+1}$.
    \ENDFOR
\ENDFOR

\STATE \textbf{return} $\mathcal{D}_{\text{sft}}$.
\end{algorithmic}
\end{algorithm}

\section{More Implementation Details}
\subsection{Cold-start Supervised Fine-Tuning (SFT).}
We initialize policy learning with a cold-start supervised fine-tuning (SFT) stage using multi-turn ReAct trajectories.
Training is conducted with \texttt{verl.trainer.fsdp\_sft\_trainer} under a fully sharded data-parallel (FSDP) setup on 8xA100 80GB GPUs.
% The SFT dataset is stored in Parquet format, where each example contains a sequence of multi-turn messages under the \texttt{messages} field, and multi-turn training is explicitly enabled.

We fine-tune the base model for 2.5 epochs with a learning rate of $5e^{-6}$ and a global training batch size of 16, using a micro-batch size of 4 per GPU.
The maximum sequence length is set to 32{,}768 tokens.
Padding tokens are removed during training to improve efficiency.
To support long-context training, we use Ulysses sequence parallelism with a parallel size of 2.

\subsection{Reinforcement Learning with GRPO.}
Starting from the cold-start SFT checkpoint, we further optimize the model using Group Relative Policy Optimization (GRPO), implemented via \texttt{verl.trainer.main\_ppo}.
The SFT checkpoint is used to initialize both the actor and the reference model.
Training is performed on 8 GPUs with a global batch size of 256 and PPO mini-batches of size 32.
The actor learning rate is set to $1e{-6}$, and training is run for 80 steps.

For each input query, the actor samples $N{=}16$ rollouts per GRPO group.
Each rollout is allowed up to 8 assistant turns, with a maximum prompt length of 16{,}384 tokens and a maximum response length of 4{,}096 tokens.
Rollouts are generated asynchronously to improve the throughput supported by the \texttt{vllm} engine.
During the rollout phase of RL training, decoding is intentionally stochastic to encourage exploration, using temperature $=1.0$, top-$p{=}1.0$, and top-$k{=}{-}1$.
Durning the rollout phase of RL training, we set temperature $=1.0$, top-$p{=}1.0$, and top-$k{=}{-}1$ to encourage the exploration. We also set Ulysses sequence parallel size as 2 and enable gradient checkpoint for long-context training.

The reward function combines format compliance and answer correctness.
We assign a format weight $\alpha=0.2$, and use a correctness threshold $\tau=0.8$ to determine whether a generated answer is considered correct.
KL regularization is not included in the reward (KL coefficient set to zero), and no critic warmup is applied.

% \paragraph{Handling Void Turns and Training Stability.}~\hansi{need to add detailed equation}
% In practice, multi-turn research agents may produce a non-trivial fraction of \emph{void turns} (approximately 10--15\% in our experiments), where a trajectory fails to generate a valid final answer.
% Such failures typically arise when the trajectory exceeds the maximum token or turn limit, or when the agent produces an invalid tool call that violates the required JSON schema.

% To improve training stability and sample efficiency, we retain these trajectories for advantage computation but exclude them from the policy loss.
% Concretely, let $\mathcal{U}$ denote the index set of trajectories with void turns within a GRPO group.
% For each $i \in \mathcal{U}$, we set the loss mask to zero for all tokens, i.e., $\{m_{i,t}\}_{t=1}^{|y_i|} = 0$, while still computing their outcome-based reward $r(q, y_i)$.
% Since these trajectories neither produce a valid answer nor follow the required format, their reward is set to $r(q, y_i) = -\alpha$.
% This design prevents malformed trajectories from directly updating the policy, while still allowing them to contribute to relative advantage estimation within the GRPO group.

\subsection{Tricks for RL training stability}~\label{appendix:trick-for-rl}

\paragraph{Masked loss for void turns}
Void turns are defined as rollouts that fail to produce a valid final answer \emph{only} because they exceed the maximum token limit or the maximum interaction-turn limit (i.e., truncation by the environment). In this case, the rollout is incomplete and does not admit a well-formed final answer.

Let $x \in \mathcal{D}$ denote a prompt, and let $\{y_i\}_{i=1}^{G}$ be a group of rollouts sampled under the previous policy $\pi_{\theta_{\text{old}}}(\cdot\mid x)$. We define the index set of void trajectories as
\begin{equation}
\mathcal{U} \triangleq \{\, i \mid y_i \text{ contains a void turn } \,\}.
\end{equation}

For each rollout $y_i$, we compute an outcome-based reward $r(q,y_i)$ following the reward definition in Section~\ref{method:rl-with-sft}. For void trajectories $i \in \mathcal{U}$, the reward is set to
\begin{equation}
r(q,y_i) = -\alpha,
\end{equation}
where $\alpha$ is the format related reward coefficient. 

\smallskip
Rather than letting void trajectories directly update the policy, we \emph{exclude the entire trajectory} from the policy loss while retaining it for \emph{group-relative advantage estimation}. Concretely, for all $i \in \mathcal{U}$, we set the token-level loss mask to zero for \emph{every} token in the rollout:
\begin{equation}
m_{i,t} = 0 \qquad \forall\, t \in \{1,\dots,|y_i|\}.
\end{equation}
For non-void trajectories $i \notin \mathcal{U}$, the mask follows our standard GRPO setting: $m_{i,t}=0$ for tokens originating from tool responses and $m_{i,t}=1$ otherwise.

\smallskip
With this trajectory-level masking, the GRPO objective becomes
\begin{equation}
\label{eq:grpo_void_mask}
\mathcal{J}_{\text{GRPO}}(\theta)
=
\mathbb{E}_{x \sim \mathcal{D},\;\{y_i\}_{i=1}^{G} \sim \pi_{\theta_{\text{old}}}(\cdot \mid x)}
\Bigg[
\frac{1}{G}\sum_{i=1}^{G}
\frac{1}{\sum_{t=1}^{|y_i|} m_{i,t}}
\sum_{t=1}^{|y_i|}
m_{i,t}\,
\min\!\Big(
\rho_{i,t}(\theta)A_{i,t},\;
\mathrm{clip}(\rho_{i,t}(\theta),1-\epsilon,1+\epsilon)A_{i,t}
\Big)
\Bigg],
\end{equation}
where $\rho_{i,t}(\theta)=\frac{\pi_{\theta}(y_{i,t}\mid c_{i,t})}{\pi_{\theta_{\text{old}}}(y_{i,t}\mid c_{i,t})}$ is the importance ratio and $A_{i,t}$ is the group-normalized relative advantage computed from the rollout-level rewards $\{r(q,y_j)\}_{j=1}^{G}$.

Because $m_{i,t}=0$ for all $t$ when $i \in \mathcal{U}$, void trajectories contribute to the group statistics used to compute $A_{i,t}$ (via their rewards), but induce \emph{zero gradient} with respect to the policy parameters. This prevents truncated rollouts from directly destabilizing policy updates, while still allowing them to shape relative advantage estimation within each GRPO group.

\paragraph{Handling JSON schema errors in tool calls.}
For tool-using agents, each tool invocation is expected to follow a predefined JSON schema.
In early implementations, when the model produced an invalid JSON output during tool calling, we returned the corresponding parsing error (e.g., raised by \texttt{json.loads}) as a tool response and allowed the generation to continue.
A typical error message is of the form: \texttt{JSONDecodeError: Expecting ',' delimiter at position 42}.

We found that this strategy works reasonably well for larger models such as Qwen3-8B.
However, for smaller models (e.g., Qwen3-4B), this behavior leads to severe training instability: after approximately 40 RL steps, the model frequently collapses into producing a large number of malformed tool calls with invalid JSON schemas, which crash the training.

To mitigate this issue, we adopt a stricter handling strategy for JSON schema violations.
Specifically, for any rollout that produces an invalid JSON schema during tool invocation, we immediately terminate the trajectory and treat it as a format violation.

\subsection{Details of Web Search Tool}
We implement the \texttt{web\_search} tool using the Google Serper API to obtain the top-10 search results
for each query.
The raw search results returned by Serper are provided in JSON format and contain
structured metadata for each URL.

Instead of exposing the raw JSON output to the LLM, we convert the search
results into a concise Markdown-style textual representation.
Each search result is formatted as:
\begin{quote}
\texttt{\{title\} (\{url\}) \{date\} \\
\{snippet\} \\
\{page\_info\}
}
\end{quote}

Here, the \texttt{snippet} is a short, search-engine-generated summary describing the
content of the corresponding webpage, and \texttt{page\_info} optionally contains
additional lightweight metadata when available.
This representation provides the model with a compact and human-readable overview of
each candidate URL before any webpage crawling is performed.

Concretely, for each query, we extract the top-10 organic search results from the
Serper JSON response.
For each result, we parse the title, URL, publication date (if available), and snippet,
and serialize them into the Markdown format described above.
The formatted entries are concatenated to form the final search result block that is
fed directly to the model as part of the web search observation.

To improve robustness, each search request is retried up to five times in the presence
of timeout or transient HTTP errors.
To reduce API cost and improve retrieval throughput, all
search results are cached at the query level, and repeated queries reuse cached results
without invoking the external search API.

\subsection{Details of Crawl Webpage Tool}

We implement the \texttt{crawl\_webpage} tool using the open-sourced
\texttt{crawl4AI}\footnote{https://github.com/unclecode/crawl4ai} library, which is used to fetch and preprocess webpage content
given a URL.
The tool is designed to preserve useful textual information from webpages while
reducing noise.
This is achieved through a set of content-filtering configurations:

First, we remove non-informative webpage components based on HTML tags, including
navigation bars, footers, sidebars, scripts, styles, and embedded media elements
(e.g., \texttt{nav}, \texttt{footer}, \texttt{script}, \texttt{style},
\texttt{iframe}, \texttt{video}).
This step eliminates layout and boilerplate content that is irrelevant to semantic
reasoning.

Second, we remove all hyperlinks and images from the extracted text.
Empirically, we find that modern LLM tokenizers are poorly suited for processing
raw hyperlinks, which often expand into pathologically long token sequences after
tokenization and significantly increase context length.
Moreover, our task is purely text-based, and images do not provide useful signals
for downstream reasoning.
Removing links and images therefore both reduces token overhead and improves the
stability of the model input.

Finally, the cleaned webpage content is converted into Markdown format.
During this process, we apply an additional pruning step provided by
\texttt{crawl4AI}, which removes extremely short text fragments with fewer than
10 words.
This further filters out low-information content and results in a compact,
LLM-friendly representation of the webpage.

\section{Case Study}
Table~\ref{appendix:case_study_1} presents a case study of \framework on the GAIA dataset. This example illustrates why webpage crawling is useful: initial web-search results only provide high-level information, and \framework clicks into a relevant page to retrieve the standardized opposite-face color pairs, which provide the key constraint needed to solve the cube reasoning problem.

Tables~\ref{appendix:case_study_2} and~\ref{appendix:case_study_3} compare the proposed model with the baseline (RL with cold-start) on the same HotpotQA question. \framework produces the correct answer by performing more search steps to collect and verify factual information, whereas the baseline in Table~\ref{appendix:case_study_3} answers prematurely before sufficient evidence is gathered.
\begin{table}[h]
    \footnotesize
    \caption{\framework case study}
    \begin{tabular}{p{13.5cm}}
        \hline
        \textbf{Question}: A standard Rubik’s cube has been broken into cubes making up its sides. The cubes are jumbled, and one is removed. There are 6 cubes with one colored face, 12 edge cubes with two colored faces, and 8 corner cubes with three colored faces. All blue cubes have been found. All cubes directly left, right, above, and below the orange center cube have been found, along with the center cube. The green corners have all been found, along with all green that borders yellow. For all orange cubes found, the opposite face’s cubes have been found. The removed cube has two colors on its faces. What are they? Answer using a comma separated list, with the colors ordered alphabetically.\\
        \hline
        \textbf{Ground Truth}: green, white \\
        \hline
        \textbf{Dataset domain}: GAIA \\
        \hline
        \textbf{\framework}: \\
        \reason{To solve this, I need the Rubik's cube piece taxonomy (6 centers / 12 edges / 8 corners) and, crucially, the standardized opposite-face color pairs (e.g., blue opposite green, orange opposite red, white opposite yellow). I'll look up the standard color arrangement.}\\
        \call{\{"name": "web\_search", "query": "standard Rubik's cube composition"\}} \\
        \response{
        [1] [All About Rubik's Cube - Cubelelo](\url{https://www.cubelelo.com/blogs/cubing/all-about-rubik-s-cube}) Structure - The Rubik's cube consists of a spherical core that holds all the pieces together, all the pieces move relative to the center core pieces. There are 20 total pieces excluding the centerpieces, which include 8 corner pieces and 12 edge pieces . 
        
         [2] [Rubik's Cube - Wikipedia](\url{https://en.wikipedia.org/wiki/Rubik\%27s_Cube}) Mechanics. A standard Rubik's Cube measures 5.6 centimetres (2+1⁄4 in) on each side. The puzzle consists of 26 unique miniature cubes, also known as "cubies" ...
         
        \centerline{$\vdots$}
         
         [10] [[PDF] Mathematics of the Rubik's cube](\url{https://www.fuw.edu.pl/~konieczn/RubikCube.pdf}) May 22, 1992 Following the standard convention, the Rubik's cube will act on the set of facets of the cube on the right. Definition 99. Let G act on a ...
        }  \\
        \hdashline
        \reason{I need more specific information about the Rubik's cube structure since the initial search didn't provide enough details about the different types of cubes. Let me search specifically for information about the different types of cubes in a Rubik's cube - the corner, edge, and center pieces - to understand their properties and how they relate to the problem. This will help me determine which cubes could be the one with two colors that was removed.}\\
        \call{\{"name": "web\_search", "query": "Rubik's cube types of cubes" \}} \\
        \response{
        [1] [10 Different Types of Rubik's Cubes and Their Unique Features](\url{https://particula-tech.com/blogs/news/10-different-types-of-rubiks-cubes}) All Types of Rubik's Cubes
        
        [2] [List of rubiks cubes and variations with pictures and names? : r/Cubers](\url{https://www.reddit.com/r/Cubers/comments/3x5iu1/list_of_rubiks_cubes_and_variations_with_pictures/}) Dec 17, 2015 I want a list of images of rubiks cubes and variations with names next to them like pyraminx, 3x3, megaminx, rubiks revenge, alexangers star, void cube, and ...

        \centerline{$\vdots$}
        
        [10] [14 Coolest and the Most Interesting Cubes of All Time - Cubelelo](\url{https://www.cubelelo.com/blogs/cubing/coolest-rubiks-cubes?srsltid=AfmBOoqP9TUNA1XCo0_6znSnIO831An3sWqtVxD0xNxTMtNz_3JhJ-D8}) Apr 15, 2025 1. Face-Turning Octahedron (FTO) · 2. Mirror Cube · 3. Megaminx · 4. Pyraminx · 5. 4x4 \& 5x5 (Rubik's Revenge \& Professor Cube) · 6. Big Cubes (8x8 ...
        } \\
        \hdashline
        \reason{I need more specific information about the Rubik's cube structure to solve this problem. The search results so far haven't given me the detailed information about the different types of cubes and their color configurations. Let me check the Wikipedia page on Rubik's Cube, which should provide comprehensive information about its structure, including how the cubes are arranged and which colors they have. This will help me understand which cubes could be the one with two colors that was removed.} \\
        \call{\{"name": "crawl\_webpage", "url": "\url{https://en.wikipedia.org/wiki/Rubik\%27s_Cube}"} \\
        \response{... Since 1988, the arrangement of colours has been standardised, with white opposite yellow, blue opposite green, and orange opposite red} \\
        \hdashline
        \textcolor{gray}{...} \\
        \textcolor{gray}{Two more web searchs (omitted for saving space)} \\
        \textcolor{gray}{...} \\
        \hdashline
        \reason{... The removed cube is an edge piece with two colors. Based on the clues, the only edge not adjacent to the orange center and not conflicting with other conditions is the **white-green** edge. Ordered alphabetically, the colors are **green, white**. ...} \\
        \ans{green, white}  \textcolor{ForestGreen}{\checkmark}\\
        \hline
    \end{tabular}~\label{appendix:case_study_1}
\end{table}

\begin{table}[h]
    \footnotesize
    \caption{\framework case study}
    \begin{tabular}{p{13.5cm}}
        \hline
        \textbf{Question}: Who directed the second film in a British series of action comedy film parodying the James Bond secret agent genre with comedy similar to Rowan Atkinson's Mr. Bean character?\\
        \hline
        \textbf{Ground Truth}: Oliver Parker \\
        \hline
        \textbf{Dataset domain}: HotpotQA \\
        \hline
        \textbf{\framework}: \\
        \reason{I need to start by identifying the British action comedy series that parodies the James Bond genre. The question mentions it has comedy similar to Rowan Atkinson's Mr. Bean character, so I'll search for information about this specific British series first. This will help me identify which film is considered the "second" in the series, and then I can find out who directed it.}\\
        \call{\{"name": "web\_search", "query": "British action comedy series parodying James Bond with Mr. Bean style comedy"\}} \\
        \response{
        [1] [What are your thoughts on the Johnny English series? The parody of ...]
            (\url{https://www.reddit.com/r/JamesBond/comments/1cqk0le/what_are_your_thoughts_on_the_johnny_english/})
            May 12, 2024.
            I prefer probably Rowan Atkinson as Johnny English than Mr Bean. The second movie is my favorite because it has the best balancing between humor and action \ldots
    
        [2] [Johnny English (film series) - Wikipedia]
            (\url{https://en.wikipedia.org/wiki/Johnny_English_(film_series)})
            Johnny English is a series of spy action comedy films parodying the James Bond secret agent genre. It features Rowan Atkinson as the title character.
    
        \centerline{$\vdots$}
    
        [10] [How would James Bond react if he encountered someone ... - Quora]
            (\href{https://www.facebook.com/lilronnymothaf/posts/-johnny-english-reborn-2011directed-by-oliver-parker-starring-rowan-atkinson-gil/1271602630992141/}
            {\url{https://www.facebook.com/.../1271602630992141}}
            ) Mr. Bean is a comedian, got nothing to do with serious businesses what the Bond in Craig era does. Johnny English series is somewhat a parody of \ldots
}

          \\
          \hdashline
          \reason{Based on the search results, I've identified that the British action comedy series parodying James Bond with Mr. Bean-style comedy is "Johnny English." The search results mention it's a series of spy action comedy films parodying the James Bond genre, featuring Rowan Atkinson who also played Mr. Bean. Now I need to find out who directed the second film in this series. Let me search specifically for information about the director of the second Johnny English movie.} \\
          \call{\{"name": "web\_search", "query": "director of second Johnny English film"\}} \\
          \response{
        [1] [Johnny English Reborn - Wikipedia](\url{https://en.wikipedia.org/wiki/Johnny_English_Reborn}) 
        Johnny English Reborn is a 2011 British spy action comedy film directed by Oliver Parker and written by Hamish McColl, based on a story by William Davies.
        
        [2] [Johnny English Reborn (2011) - IMDb](\url{https://www.imdb.com/title/tt1634122/}) 
        IMDb entry for \emph{Johnny English Reborn}. The film is directed by Oliver Parker and stars Rowan Atkinson, featuring a spy parody narrative in which Johnny English confronts international threats.
        
        \centerline{$\vdots$}

        [10] [Johnny English Reborn (2011) Directed by: Oliver Parker Starring ...](\href{https://www.facebook.com/lilronnymothaf/posts/-johnny-english-reborn-2011directed-by-oliver-parker-starring-rowan-atkinson-gil/1271602630992141/}
{\url{https://www.facebook.com/.../1271602630992141}}) 
        Jun 25, 2025. A Facebook post summarizing \emph{Johnny English Reborn} (2011), confirming Oliver Parker as the director and listing the main cast including Rowan Atkinson, Gillian Anderson, Rosamund Pike, and Dominic West.
        }

        \\
        \hdashline

        \reason{Based on the search results, I've identified that the British action comedy series parodying James Bond with Mr. Bean-style comedy is "Johnny English." I've also found that there were multiple films in this series, with the second one directed by Oliver Parker. However, I should verify this information and get more details about the series and its films. Let me search for more specific information about the Johnny English film series to confirm who directed the second film.} \\
        \call{\{"name": "web\_search", "query": "Johnny English film series director"\}} \\
        \response{
        [1] [Johnny English - Wikipedia](\url{https://en.wikipedia.org/wiki/Johnny_English}) 
        Johnny English is a 2003 British spy action comedy film directed by Peter Howitt and written by Neal Purvis, Robert Wade, and William Davies.
        
        [2] [Johnny English (film series) - Wikipedia](\url{https://en.wikipedia.org/wiki/Johnny_English_(film_series)}) 
        Overview of the \emph{Johnny English} film series. The franchise parodies the James Bond secret agent genre and features multiple directors including Peter Howitt, Oliver Parker, and David Kerr.
        
        \centerline{$\vdots$}
        
        [10] [Johnny English Collection — The Movie Database (TMDB)](\url{https://www.themoviedb.org/collection/86117-johnny-english-collection?language=en-US}) 
        TMDB collection page summarizing the \emph{Johnny English} films, describing the series as British action comedies parodying the James Bond genre and starring Rowan Atkinson as the titular character.
        }

        \\
        \hdashline

        \textcolor{gray}{...} \\
        \textcolor{gray}{One more web searches (omitted for saving space)} \\
        \textcolor{gray}{...} \\

        \hdashline
        
        \reason{The British action comedy spy parody series refers to the Johnny English films starring Rowan Atkinson. 
The second film in the series is Johnny English Reborn (2011), which was directed by Oliver Parker.
...} \\
        \ans{Oliver Parker} \textcolor{ForestGreen}{\checkmark} \\
        \hline
    \end{tabular}~\label{appendix:case_study_2}
\end{table}

\begin{table}[h]
    \footnotesize
    \caption{RL with naive cold-start SFT case study}
    \begin{tabular}{p{13.5cm}}
        \hline
        \textbf{Question}: Who directed the second film in a British series of action comedy film parodying the James Bond secret agent genre with comedy similar to Rowan Atkinson's Mr. Bean character?\\
        \hline
        \textbf{Ground Truth}: Oliver Parker \\
        \hline
        \textbf{Dataset domain}: HotpotQA \\
        \hline
        \textbf{\framework}: \\
        \reason{Okay, let's try to figure out who directed the second film in that British action comedy parodying James Bond. The user mentioned it's similar to Rowan Atkinson's Mr. Bean, so I need to recall or find the name of that series. First, I remember that there was a British film called "The Fast Show" but I'm not sure if that's the one. Wait, maybe it's "The Nude Bomb"? No, that's a different genre. Oh, right! There's a film called "Johnny English" which is a comedy parody of James Bond. Let me confirm that. Johnny English was released in 1999, and the sequel is "Johnny English Strikes Back" in 2000 ...}\\
        \call{\{"name": "web\_search", "query": "British James Bond parody action comedy series"\}} \\

        \response{ [1] [List of James Bond parodies and spin-offs](\url{https://en.wikipedia.org/wiki/List_of_James_Bond_parodies_and_spin-offs}) 
        A Wikipedia list summarizing films and television works that parody or spin off from the James Bond franchise, including British spy comedies such as Hot Enough for June (1964) and Carry On Spying (1964).

         [2] [Parodies \& Imitations | James Bond Wiki - Fandom](\url{https://jamesbond.fandom.com/wiki/Parodies_\%26_Imitations}) 
        An overview of notable James Bond parodies and imitations, highlighting well-known examples such as the \emph{Austin Powers} film series and its parody of Bond characters and tropes.
        
        \centerline{$\vdots$}
        
        [10] [Spy Comedy Films](\url{https://www.imdb.com/list/ls040234091/}) 
        An IMDb list of spy comedy films, including entries from the \emph{Johnny English} series starring Rowan Atkinson, such as \emph{Johnny English} (2003), \emph{Johnny English Reborn} (2011), and \emph{Johnny English Strikes Again} (2018).}

          \\
          \hdashline
          
          \reason{Okay, so the user is asking about the director of the second film in a British series that parodies the James Bond genre with comedy similar to Rowan Atkinson's Mr. Bean. From the web search results, I see that the Johnny English series is mentioned as a spy action comedy parodying James Bond, and Rowan Atkinson is the main character. That must be the series they're referring to. Now, the first result mentions the List of James Bond parodies, and the fourth result is the Wikipedia page for Johnny English (film series). Let me check that. The Wikipedia page says Johnny English has two main films: the first was released in 1999, and the second, "Johnny English Strikes Back," in 2000. Next, I need to find out who directed the second film. The initial search didn't mention the director, so I should perform another web search specifically for the director of "Johnny English Strikes Back." Let me use the web\_search tool again with that query.} \\
          \call{\{"name": "web\_search", "query": "director of Johnny English Strikes Back"\}} \\
       \response{
        [1] [Johnny English Strikes Again (2018) - IMDb](\url{https://www.imdb.com/title/tt6921996/}) 
        IMDb page for \emph{Johnny English Strikes Again} (2018). The film is directed by David Kerr and stars Rowan Atkinson, continuing the British spy action comedy parody series.
        
        [2] [Johnny English Strikes Again (2018) - Full cast \& crew - IMDb](\url{https://www.imdb.com/title/tt6921996/fullcredits/}) 
        Full credits listing for the film, confirming David Kerr as the director and William Davies as the screenwriter, along with detailed production information.
        
        \centerline{$\vdots$}
        
        [10] [David Kerr on Rowan Atkinson, Johnny English Strikes Again, and Comedy](\url{https://weliveentertainment.com/welivefilm/david-kerr-on-rowan-atkinson-johnny-english-and-comedy/}) 
        An interview article published on Nov 6, 2018, featuring director David Kerr discussing his work on \emph{Johnny English Strikes Again}, his collaboration with Rowan Atkinson, and his background in comedy.
        }

        \\
        \hdashline

        \reason{Okay, so the user is asking about the director of the second film in the British series that parodies the James Bond genre, specifically mentioning Rowan Atkinson's Mr. Bean style comedy. From the previous search, we identified that the series in question is "Johnny English." The first search confirmed that "Johnny English Strikes Back" is the second film in the series. Looking at the current search results for the director of "Johnny English Strikes Back," multiple sources mention David Kerr as the director. For example, the IMDb entry lists David Kerr as the director, and other reviews and articles also confirm this. The second film, "Strikes Back," was released in 2000, and the director listed across various platforms is consistently David Kerr. I should verify if there's any conflicting information, but all the sources here align. Therefore, the answer should be David Kerr.}

        \ans{David Kerr} \textcolor{Red}{$\times$} \\
        \hline
    \end{tabular}~\label{appendix:case_study_3}
\end{table}

% You can have as much text here as you want. The main body must be at most $8$ pages long.
% For the final version, one more page can be added.
% If you want, you can use an appendix like this one.  

% The $\mathtt{\backslash onecolumn}$ command above can be kept in place if you prefer a one-column appendix, or can be removed if you prefer a two-column appendix.  Apart from this possible change, the style (font size, spacing, margins, page numbering, etc.) should be kept the same as the main body.
%%%%%%%%%%%%%%%%%%%%%%%%%%%%%%%%%%%%%%%%%%%%%%%%%%%%%%%%%%%%%%%%%%%%%%%%%%%%%%%
%%%%%%%%%%%%%%%%%%%%%%%%%%%%%%%%%%%%%%%%%%%%%%%%%%%%%%%%%%%%%%%%%%%%%%%%%%%%%%%

\end{document}